\documentclass[11pt]{article}

\usepackage[final]{acl}

\usepackage{times}
\usepackage{latexsym}

\usepackage[T1]{fontenc}

\usepackage[utf8]{inputenc}

\usepackage{microtype}

\usepackage{inconsolata}

\usepackage{graphicx}

\usepackage{booktabs}
\usepackage{amssymb}
\usepackage{makecell}
\usepackage{subcaption}
\usepackage{tcolorbox}
\usepackage{dblfloatfix}
\title{\textsc{ShishuLM}: Achieving Optimal and Efficient Parameterization with Low Attention Transformer Models}

\author{
  \textbf{Shivanshu Kumar\textsuperscript{1}} 
  \textbf{Gopalakrishnan Srinivasan\textsuperscript{1}}
\\
\\
  \textsuperscript{1}Department of Computer Science and Engineering, \\ Indian Institute of Technology, Madras.
\\
 \small{
   \texttt{cs23b058@smail.iitm.ac.in} \quad
   \texttt{sgopal@cse.iitm.ac.in}
 }
}

\begin{document}
\maketitle

\begin{abstract}
While the transformer architecture has achieved state-of-the-art performance on natural language processing tasks, these models impose substantial memory and computational overhead. Recent research has identified significant architectural redundancies within these models, particularly in the attention sub-layers in the top layers, presenting opportunities for optimization without compromising performance. Taking insights from research on inference-time layer pruning and depth-dependent computation in language models, we introduce an efficient language model architecture referred to as \textsc{ShishuLM}. By replacing full decoder layers at the top of the model with MLP-only blocks, we achieve up to  $10-60$\% improvement in generation latency and $1.3 -5\times$ gain in throughput. Upon further sharing parameters across adjacent MLP-only layers of \textsc{ShishuLM}, we obtain up to 20\% savings in memory with minimal degradation in performance. Our findings provide insights towards building more efficient language modeling architectures from a pre-training standpoint by leveraging how information flows in transformers.
\end{abstract}

\begin{figure*}[t]
\centering
\begin{subfigure}{0.4\linewidth}
    \includegraphics[width=\linewidth]{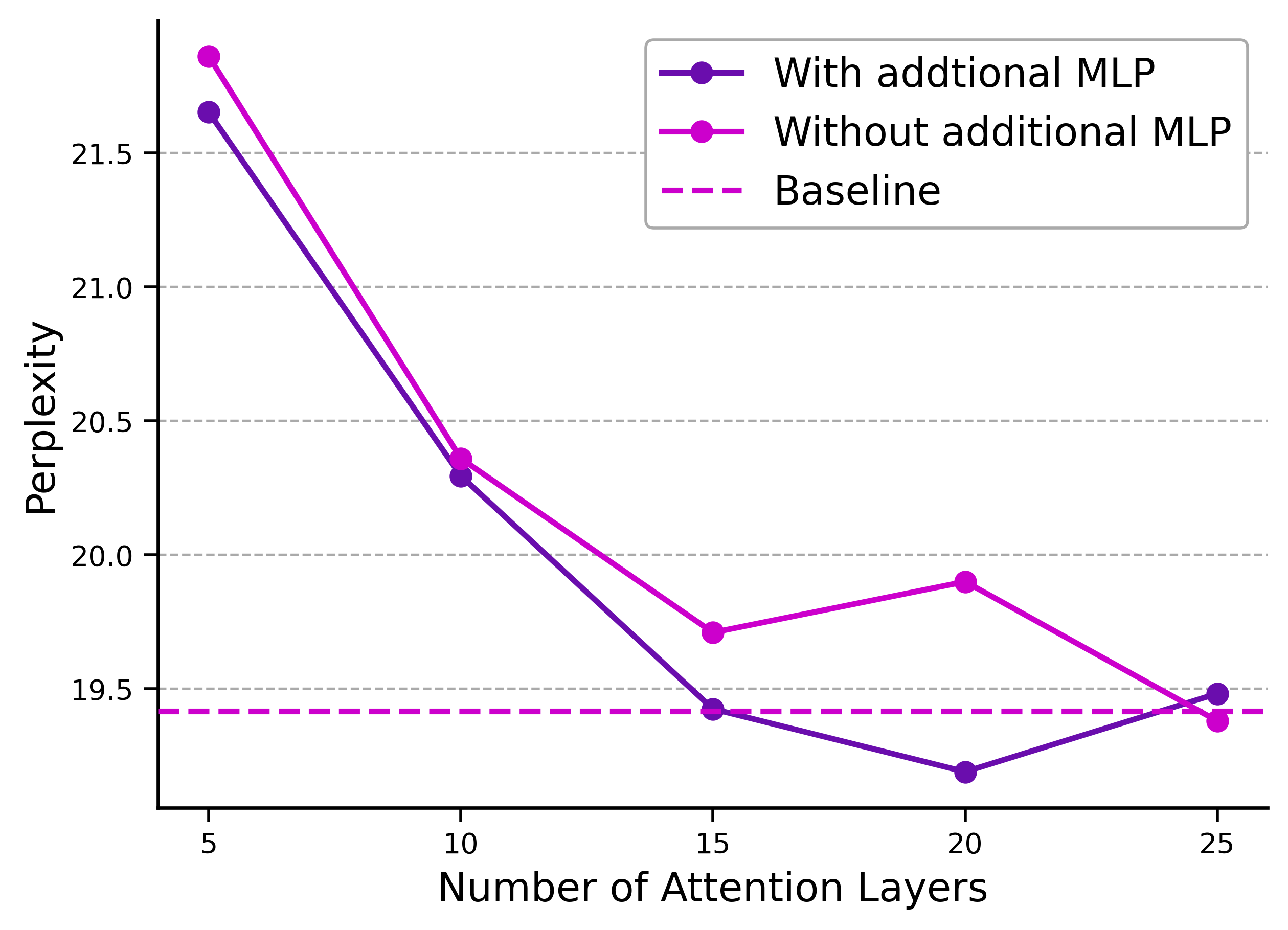}
    \caption{125M}
\end{subfigure}
\begin{subfigure}{0.4\linewidth}
    \includegraphics[width=\linewidth]{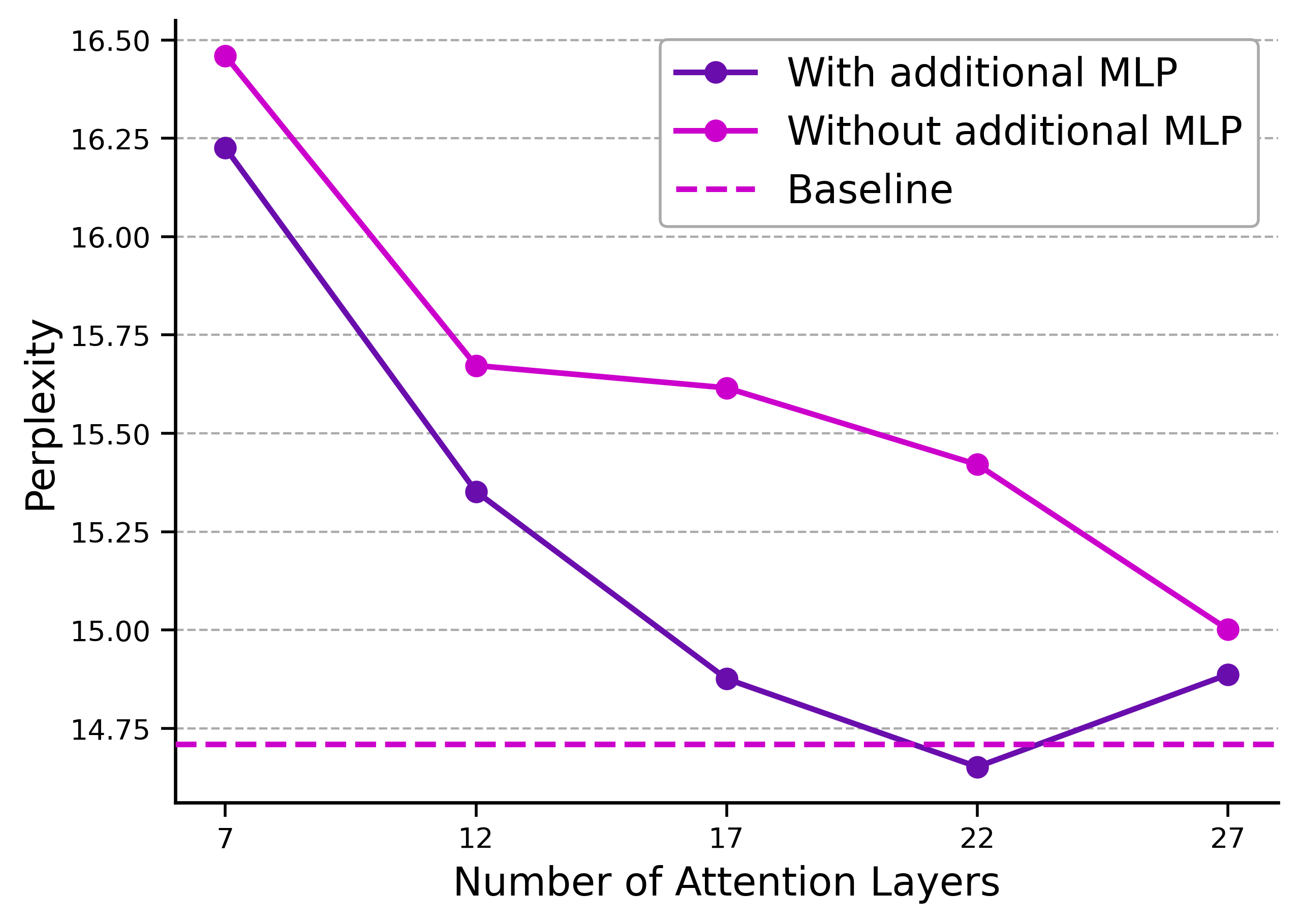}
    \caption{350M}
\end{subfigure}

\caption{Effect of removing attention layers and adding extra MLP layers to regain performance. Models in (a) and (b) are trained on 5B and 10B tokens, respectively.}
\label{fig:with-without-mlp-small}
\end{figure*}

\begin{figure*}[t]
\centering
\begin{subfigure}{0.4\linewidth}
    \includegraphics[width=\linewidth]{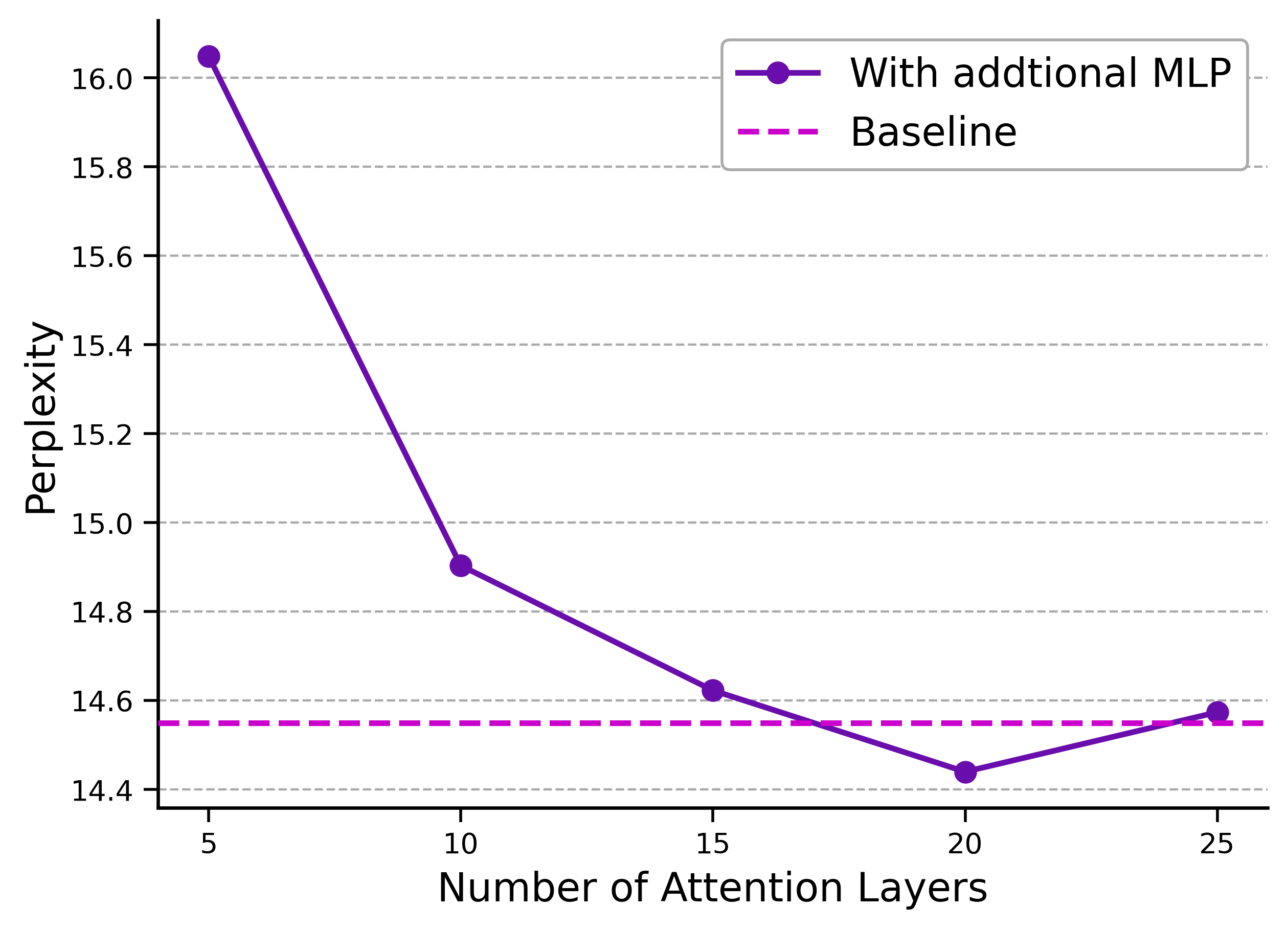}
    \caption{125M}
\end{subfigure}
\begin{subfigure}{0.4\linewidth}
    \includegraphics[width=\linewidth]{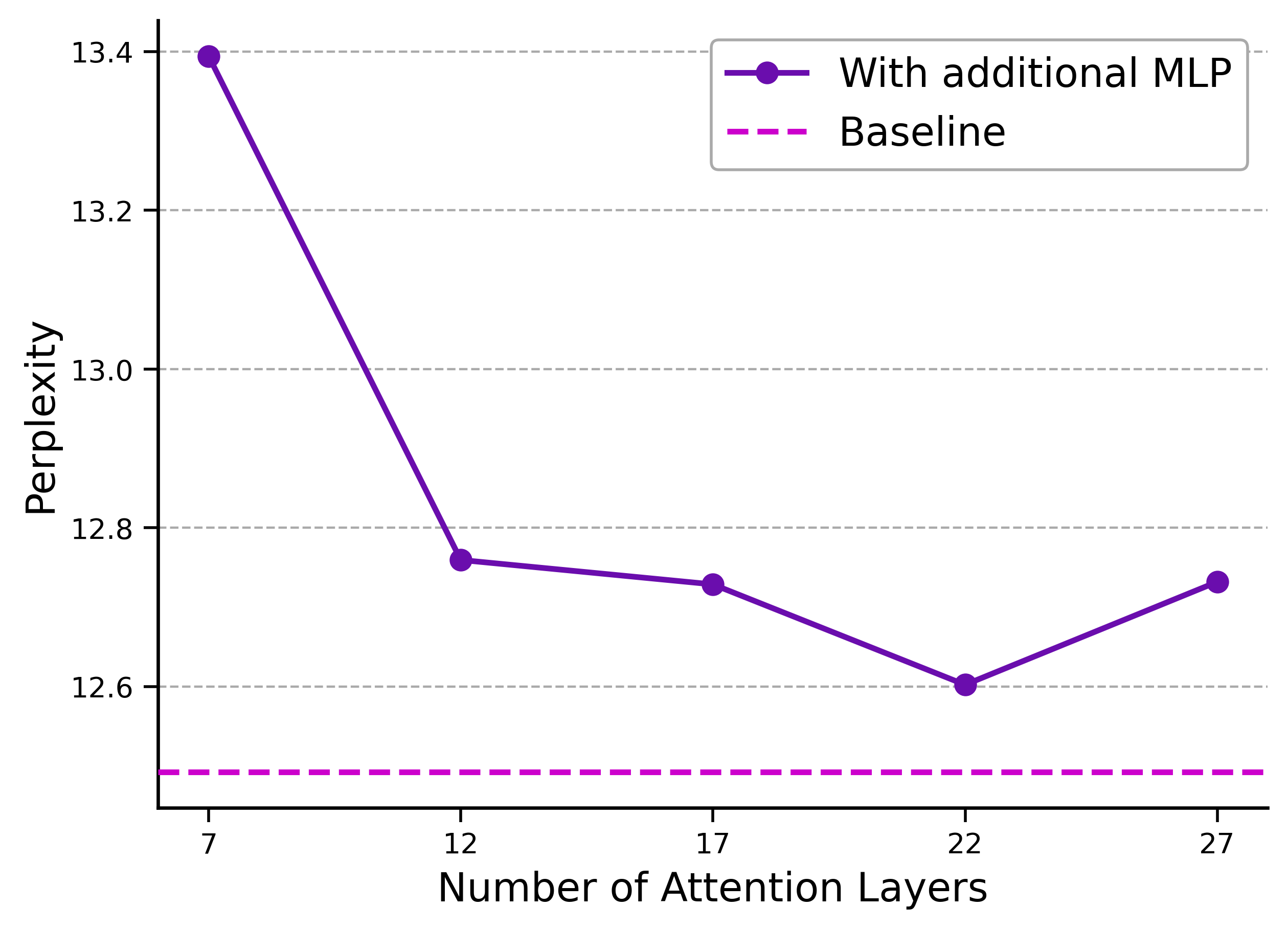}
    \caption{350M}
\end{subfigure}

\caption{Effect of removing attention layers and adding extra MLP layers, trained on 50B tokens.}
\label{fig:with-mlp-50b}
\end{figure*} 
\begin{figure*}[t]
\centering

\includegraphics[width=0.7\linewidth]{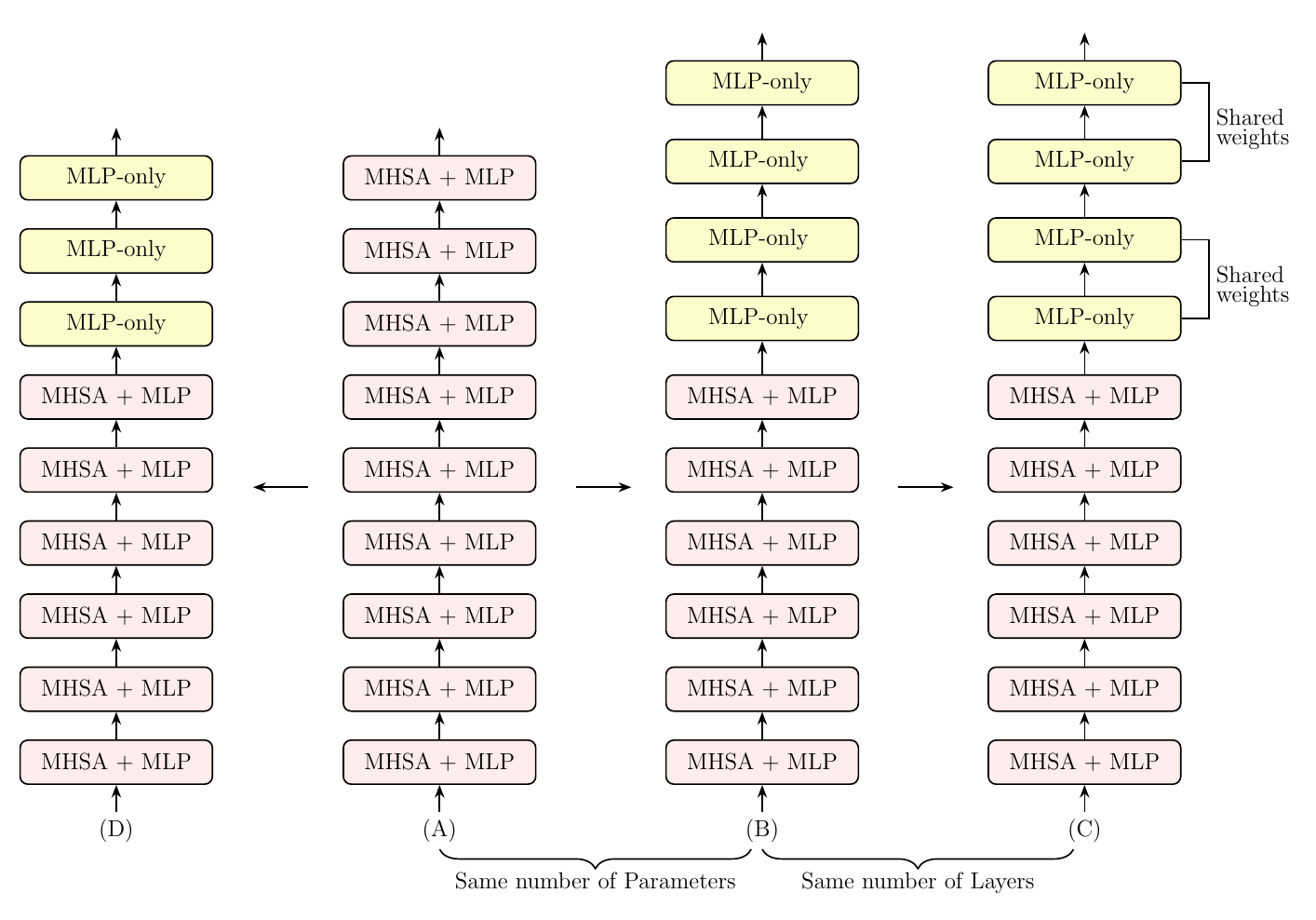}
\caption{This figure describes the architectures on which we perform our pre-training experiments. (A) shows Vanilla Transformer model with each layer as a full-decoder layer (MHSA + MLP), (D) is the result of removing MHSA from the top layers. (B) is obtained by removing attention layers from (A) and then adding extra MLPs to balance out parameters. (C) is obtained from (B) by tying together the weights of the MLP-only layers at the top, in pairs of two.}
\label{fig:architecture}
\end{figure*} 

\section{Introduction}

The transformer model architecture introduced by~\citet{attention2017vaswani} has emerged as the dominant paradigm for language modeling, achieving state-of-the-art performance across a broad range of natural language processing tasks. Decoder-only transformers consist of a stack of identical layers (blocks), each comprising a multi-head self-attention (MHSA) sub-layer followed by a position-wise feed-forward (MLP) sub-layer, with residual connections and layer normalization applied throughout. In the self-attention mechanism, each token attends over all preceding tokens by computing queries, keys, and values from the hidden representation, enabling rich contextual interactions across the sequence. The MLP sub-layer then applies a non-linear transformation independently at each token position, modulating the resulting representations.

Despite their impressive capabilities, Large Language Models (LLMs) typically contain billions of trainable parameters~\cite{bai2023qwentechnicalreport, touvron2023llamaopenefficientfoundation, jiang2023mistral7b, javaheripi2023phi}, imposing substantial memory and compute requirements. The sheer parameter count necessitates significant memory for storage alone. During training, this is compounded by the need to retain intermediate activations for backpropagation. During inference, the attention mechanism requires maintaining a key-value (KV) cache that grows linearly with both sequence length and the number of attention layers, further increasing memory pressure.

However, several lines of research have identified significant redundancies. At inference time, structured and unstructured pruning methods remove superfluous parameters from pre-trained models~\cite{ma2023llmprunerstructuralpruninglarge, frantar2023sparsegptmassivelanguagemodels}, while layer-level pruning identifies and eliminates entire decoder layers using metrics such as cosine similarity~\cite{men2024shortgpt, gromov2025the}, Shapley values~\cite{siddiqui2024deeperlookdepthpruning}, or perplexity on a calibration set~\cite{song2024sleb}. 
However, these approaches inherit the full pre-training costs of the original architecture and incur additional overhead for identifying components to prune and performance recovery through fine-tuning.

A separate line of research focuses on reducing the Key-Value (KV) cache requirements by pre-training models with the key-value pairs shared across different layers. For instance,~\citeauthor{wu2024layercondensedkvcacheefficient} compute keys and values only at a later layer and share them across bottom layers, whereas \citeauthor{sun2024cacheoncedecoderdecoderarchitectures} compute key-value pairs only for the first half of the layers and share them across the remaining layers. 

A growing body of evidence suggests that these costs are not fully justified by the computational
necessity of each architectural component. For instance,~\cite{men2024shortgpt} and~\citet{gromov2025the} both observe that later decoder blocks in LLaMA-style architectures exhibit high cosine similarity between their block inputs and outputs, suggesting near-identity transformations in those layers. 

More directly relevant to our work, studies on the \emph{no-op} behaviour of attention~\cite{bodarenko, barbero2025why} show that, in many heads, the attention pattern collapses to attending almost exclusively to a single token (often the first), effectively making the entire operation a pass-through. This is consistent with the observation of \emph{attention sinks}~\cite{xiao2024efficient}, where certain semantically meaningless tokens absorb disproportionate attention.

Furthermore, research in AI interpretability has emphasized \emph{depth-wise information flow} in transformers. Early layers primarily perform \emph{token mixing}, integrating contextual information across the sequence, while later layers largely \emph{refine the output probability distribution}, sharpening predictions~\cite{lad2024infstage, csordas2025depth, queipo2026attention}.

Building on these insights, we ask the following question:
\begin{quote}
Is there a more optimal allocation of parameters that is both high performing and efficient?
\end{quote}
 We answer this in the affirmative. Our experiments reveal that pre-training decoder-only transformers by replacing full decoder layers at the top with MLP-only sub-layers performs comparably to the full decoder  baseline at the same parameter scale, while achieving reductions in training and inference costs. Furthermore, leveraging high weight distribution similarity between adjacent MLP-only layers, we share weights across these layers in pairs, achieving further memory savings with minimal performance loss. This work is, to the best of our knowledge, among the first to directly translate insights from interpretability research into an \emph{architectural design choice at pre-training time}. We use the MobileLLM model family~\cite{mobilellm}, with parameter counts ranging from 125M to 1.5B, to validate the proposed architecture.

\paragraph{Key contributions.}
\begin{enumerate}
    \item We propose \textsc{ShishuLM}\footnote{The word \textit{Shishu} in Sanskrit means ``baby.''}, an efficient architecture for pre-training decoder-only transformer-based language models that employs only MLP sub-layers in the later blocks of the model, motivated by depth-wise redundancy in attention computation.
    \item We validate \textsc{ShishuLM} by demonstrating comparable downstream performance and improved efficiency relative to the MobileLLM baselines at scales ranging from 125M to 1.5B parameters.
    \item We further present a \emph{weight-shared} variant of \textsc{ShishuLM}, in which parameters are shared across adjacent MLP-only layers, achieving substantial memory savings with negligible performance degradation.
\end{enumerate}

\section{Preliminaries}
\subsection{Baseline architecture}

We use decoder-only transformer models as our baselines. These models consist of identical decoder layers that are stacked sequentially. Each of the decoder layers further consists of a Mutli-Headed-Self-Attention (MHSA) sub-layer followed by a Multi Layer Perceptron (MLP) sub-layer. The MHSA and MLP layers are connected in a residual fashion, i.e., the inputs are added to the outputs before being passed to the next layer. We study pre-LayerNorm architectures in this work, where the inputs are normalized before passing to each of the MHSA and MLP sub-layers. Let $L$ denote the number of decoder layers in the model, and $x_i$ be the input to the $i^{th}$ block. The forward computation is given by
$$x_{i+1} = y_i + \textrm{\ MLP\ (\ LN\ (} y_i )\ )$$
where, $$y_i = x_i +  \textrm{MHSA\ (\ LN\ (}x_i)\ ).$$
$\textrm{MLP(.), LN(.), and Self-Attention(.)}$ represent the functions computed by MLP, normalization and self-attention layers, respectively.

\subsection{Depth-wise Information Flow in Transformers}

A growing body of interpretability research has established that computation in transformer networks is not uniform across depth, but rather proceeds in functionally distinct \emph{stages}~\cite{lad2024infstage, queipo2026attention}. 
\citeauthor{lad2024infstage} identify that inference in LLMs occurs in four distinct, depth-dependent stages: (1) detokenization, (2) feature engineering, (3) prediction ensembling, and (4)
residual sharpening. Crucially, they observe that the models exhibit remarkable robustness to interventions such as layer skipping and swapping in intermediate layers. 
\citeauthor{queipo2026attention} propose a \emph{Mix-Compress-Refine framework}: Early layers (0–20\% depth) mix information broadly via diffuse attention. Middle layers (20–85\%)
compress representations while halting mixing through attention sinks and later layers (85–100\%) selectively refine through localized attention

\citeauthor{csordas2025depth} provide complementary evidence by comparing representations across models of different sizes, finding that layers at similar \emph{relative} depths map most closely to one another across architectures. This suggests that larger models do not learn qualitatively new types of computation, but rather distribute the same functional stages more finely across more layers. They further observe that later half of the model spends a disproportionate fraction of model capacity performing distribution matching rather than integrating new information and composition, characterizing this as a potentially wasteful use of parameters. 

\citeauthor{artzy-schwartz-2024-attend} reveal a depth-dependent division of attention mechanisms, showing that early layers primarily use attention for gathering information from previous tokens, while the later layers shift towards internal consolidation with significantly reduced attention importance. Their findings indicate that manipulating representations in the top 30-50\% of layers has minimal performance impact, supporting the hypothesis that layers serve fundamentally different computational purposes with changing depth.

Together, these findings motivate a simple design principle: attention is crucial in the early layers where information integration across tokens is the primary computational task, but might become redundant in later layers since they mostly perform refinement of the output distribution. We therefore attempt to explore if pre-training can be done without these sub-layers in the first place. \textsc{ShishuLM} is a direct embodiment of this principle.
\begin{figure*}
\centering

  \includegraphics[width=0.7\linewidth]{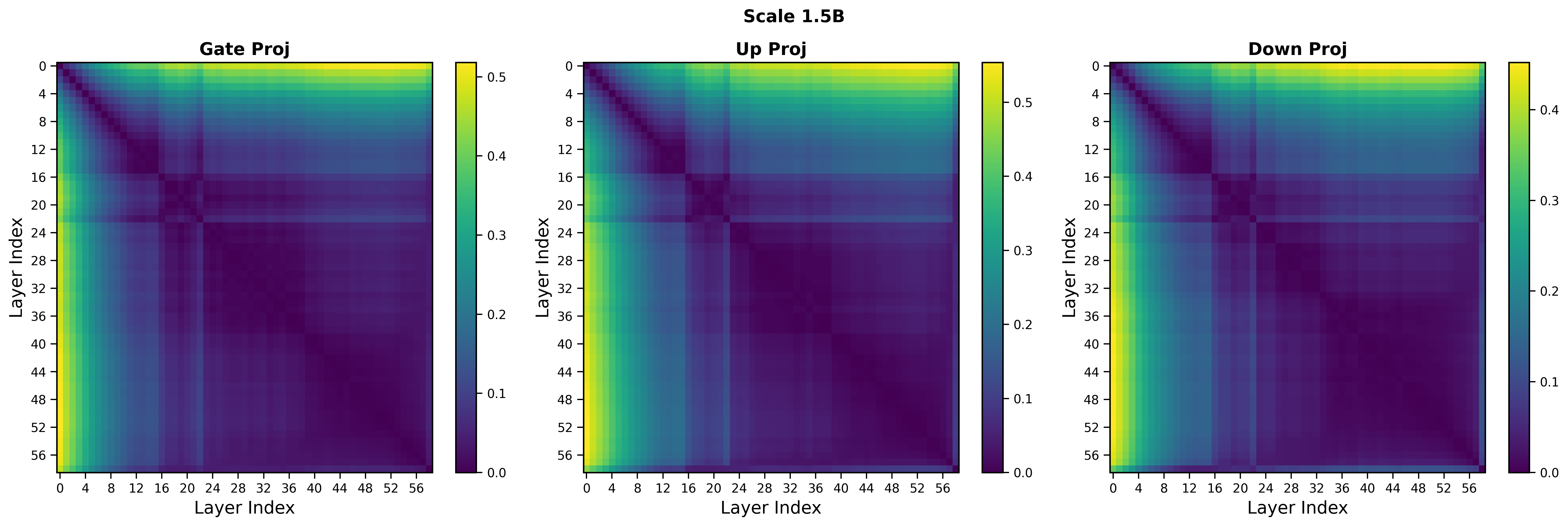}
\caption{Earth-Mover's distance across all pairs of layers in \textsc{ShishuLM-1.5B}.}
\label{fig:emd_sim}
\end{figure*}

\begin{figure*}
\centering
    \includegraphics[width=0.7\linewidth]{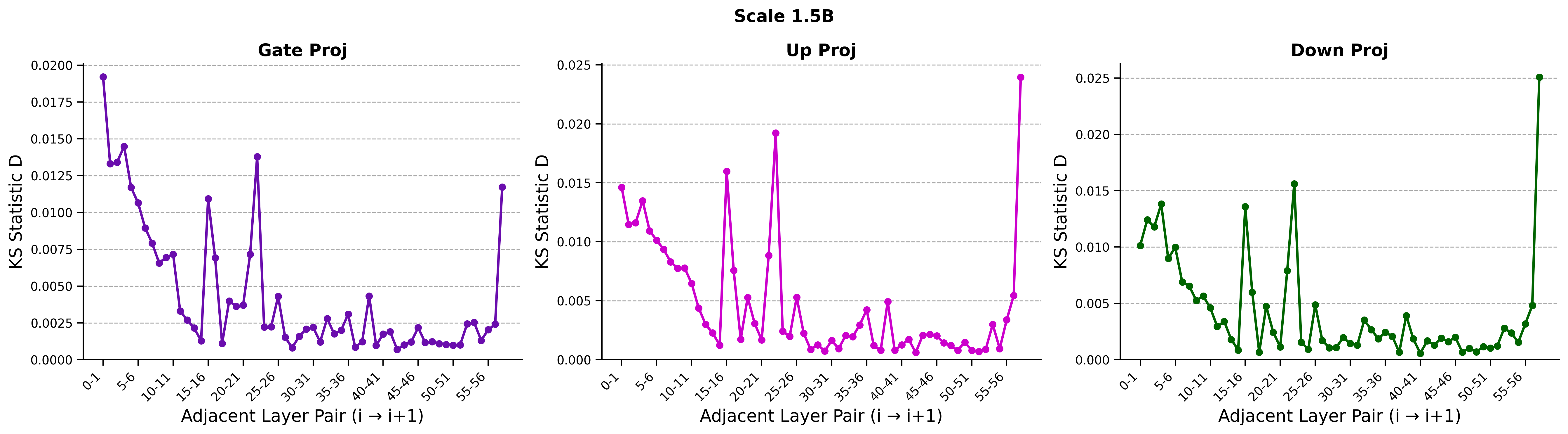}
%
\caption{KS-similarity between adjacent layers in \textsc{ShishuLM-1.5B}.}
\label{fig:ks_sim}
\end{figure*} 

\begin{figure*}[t]
\centering
\begin{subfigure}{0.4\linewidth}
    \includegraphics[width=\linewidth]{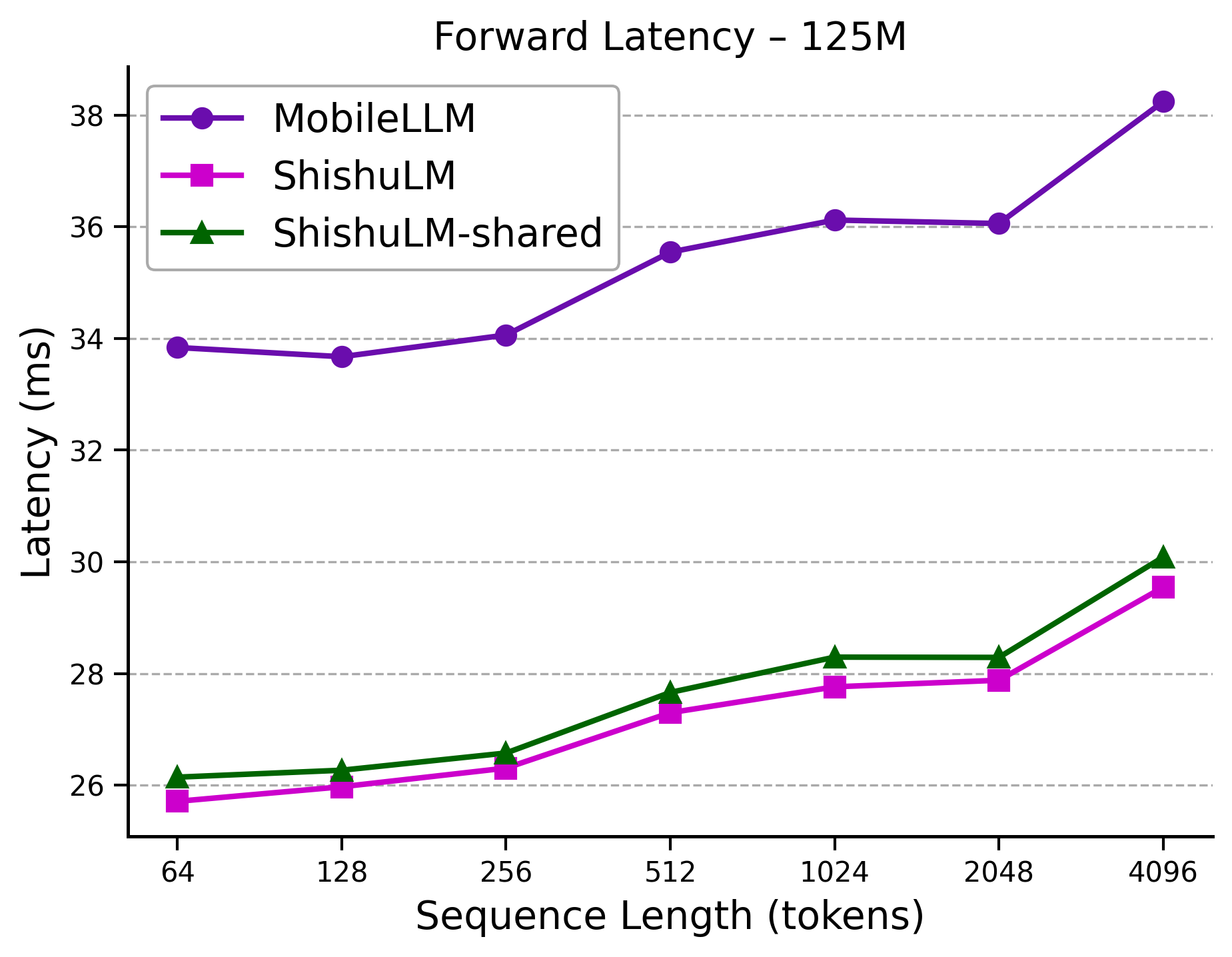}
    \caption{125M}
\end{subfigure}
\begin{subfigure}{0.4\linewidth}
    \includegraphics[width=\linewidth]{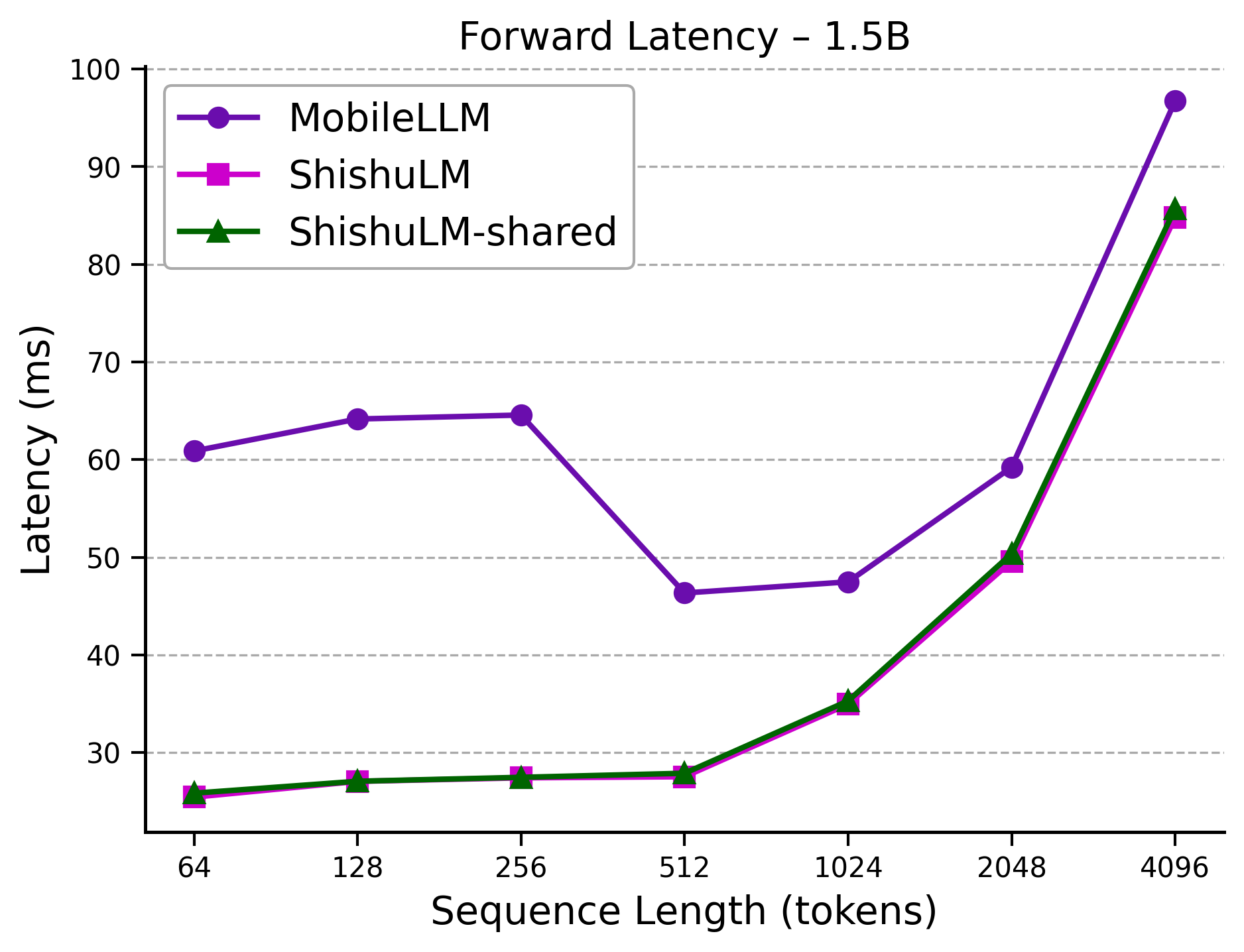}
    \caption{1.5B}
\end{subfigure}

\caption{Forward pass latency (ms) of models at the (a) 125M and (b) 1.5B scale.}
\label{fig:fwd_latency}
\end{figure*}

\begin{figure*}[t]
\centering
\begin{subfigure}{0.4\linewidth}
    \includegraphics[width=\linewidth]{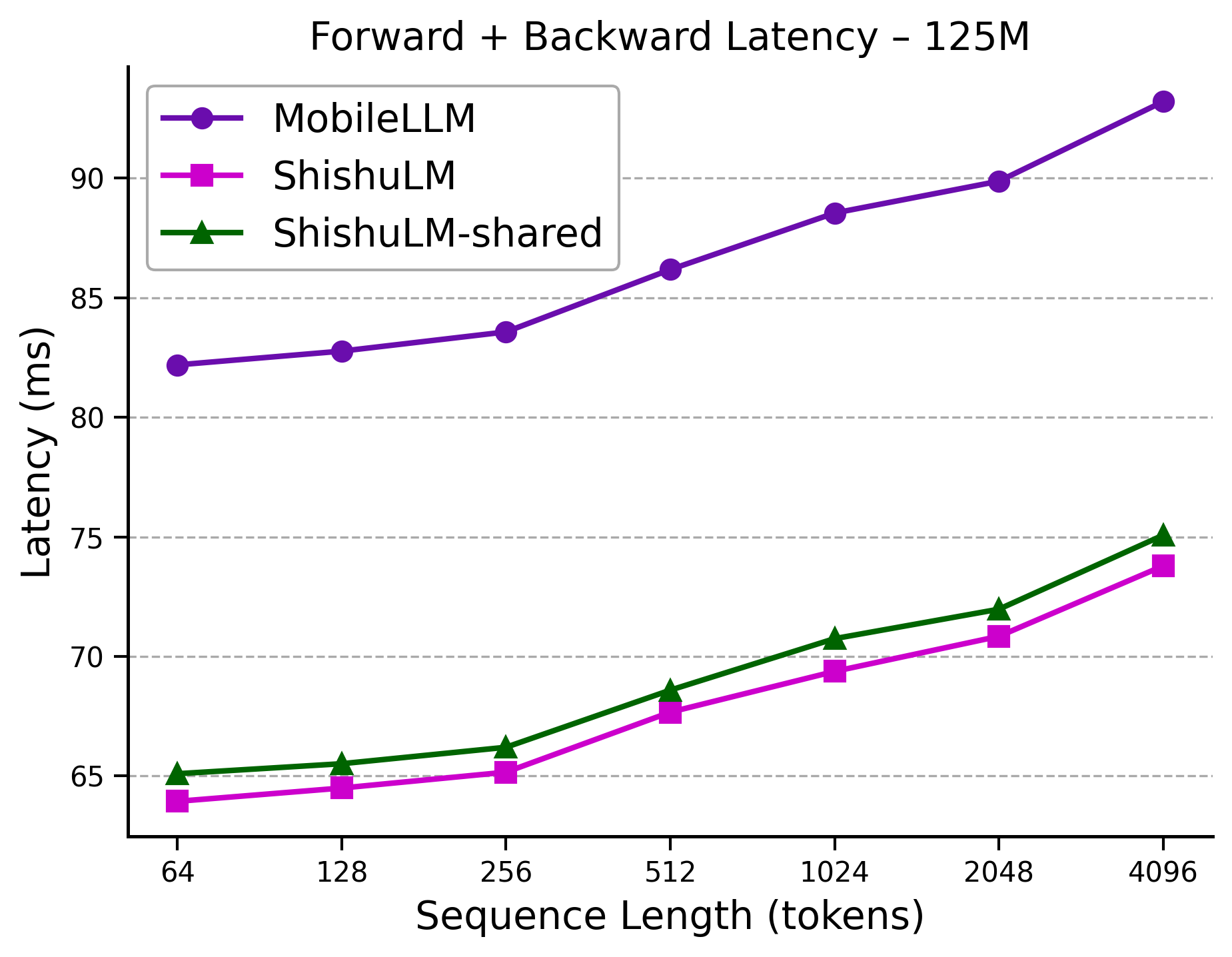}
    \caption{125M}
\end{subfigure}%
\begin{subfigure}{0.4\linewidth}
    \includegraphics[width=\linewidth]{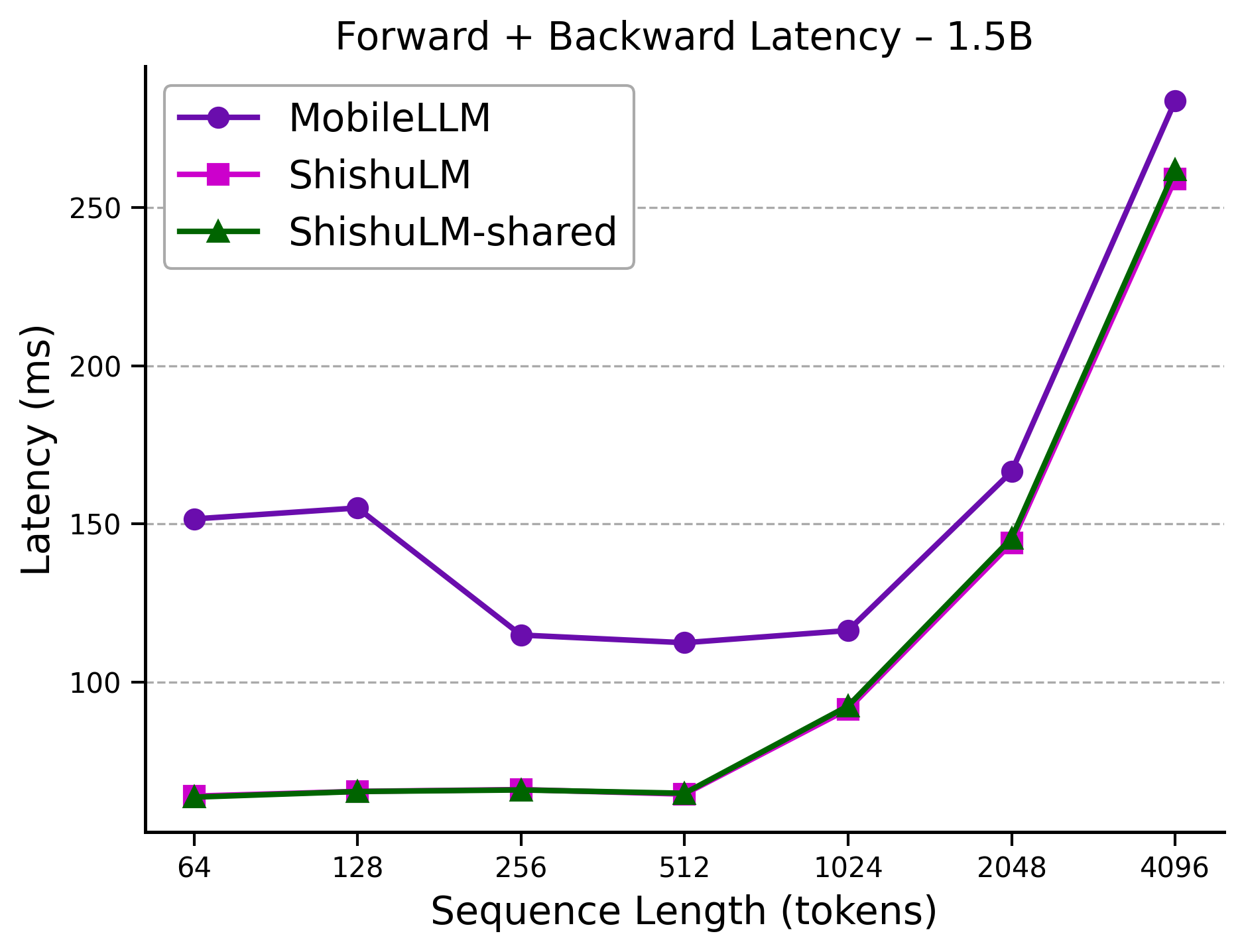}
    \caption{1.5B}
\end{subfigure}

\caption{Forward + backward pass latency (ms) of models at the (a) 125M and (b) 1.5B scale.}
\label{fig:fwd_bwd_latency}
\end{figure*}

\begin{figure*}[t]
\centering
\begin{subfigure}{0.4\linewidth}
    \includegraphics[width=\linewidth]{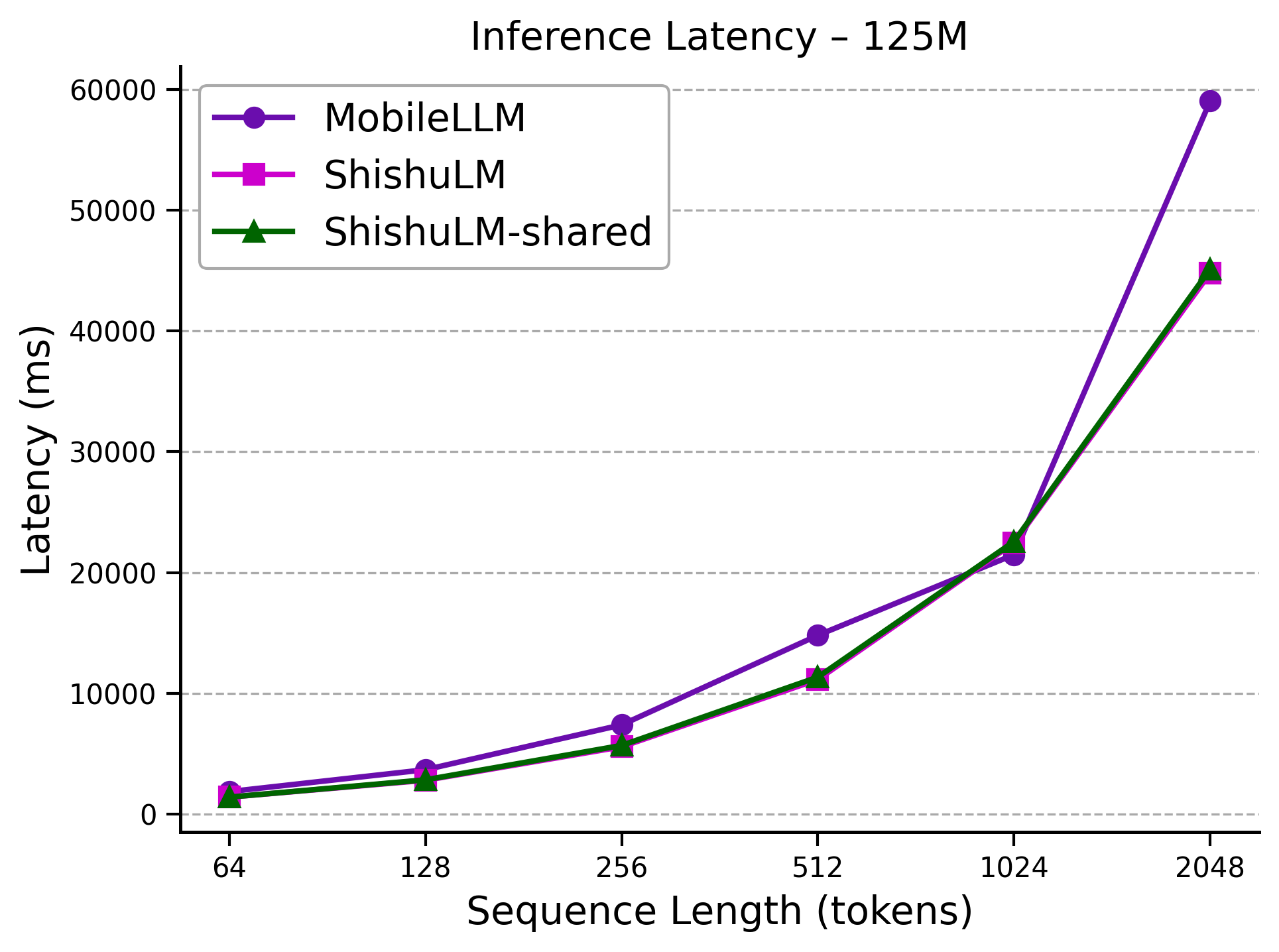}
    \caption{125M}
\end{subfigure}
\begin{subfigure}{0.4\linewidth}
    \includegraphics[width=\linewidth]{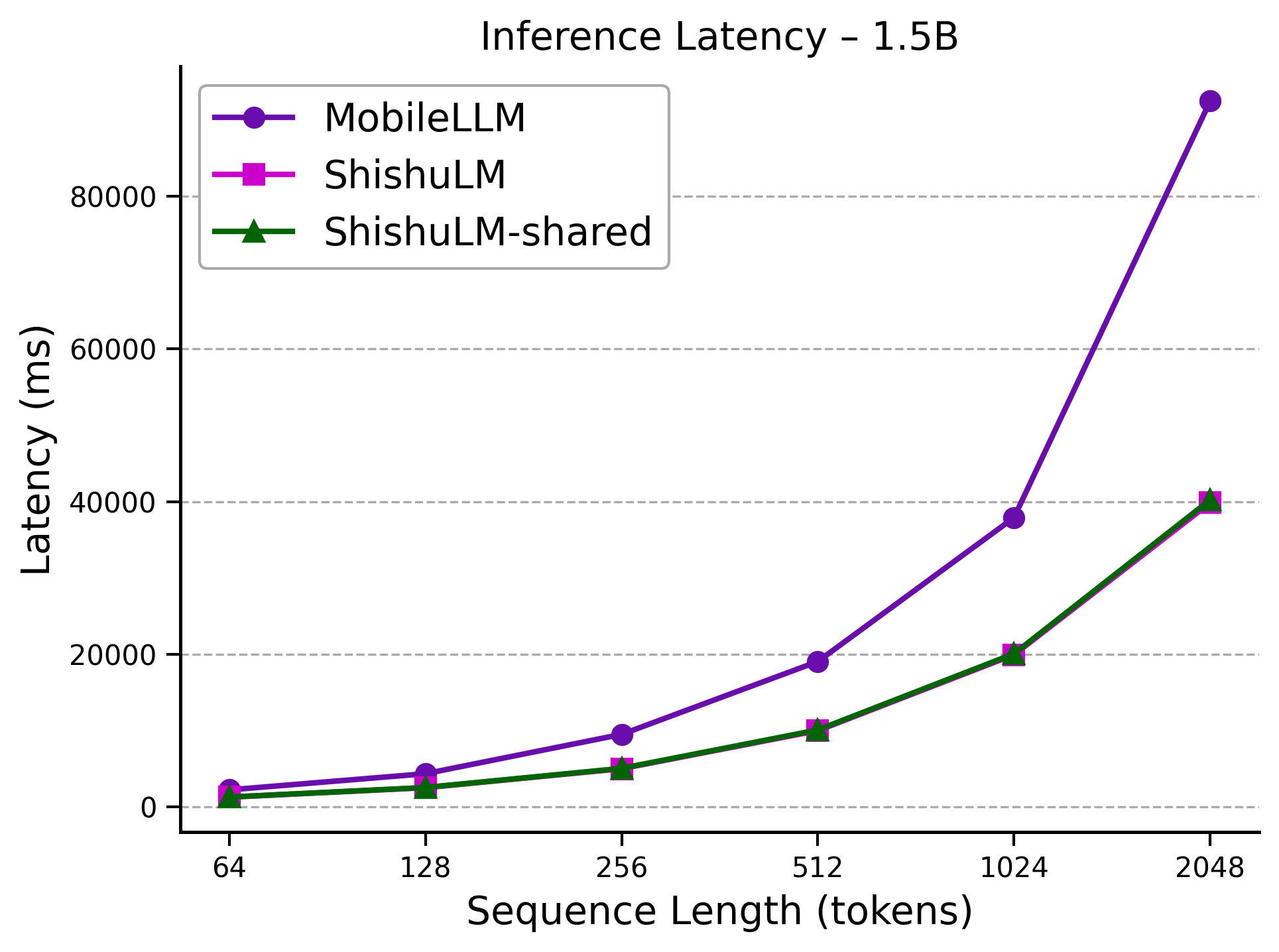}
    \caption{1.5B}
\end{subfigure}

\caption{Generation latency (ms) of models at the (a) 125M and (b) 1.5B scale, with the number of new tokens generated. Prefill Cache size is set to 2048 tokens.}
\label{fig:gen_latency}
\end{figure*}

\begin{figure*}[t]
\centering
\begin{subfigure}{0.4\linewidth}
    \includegraphics[width=\linewidth]{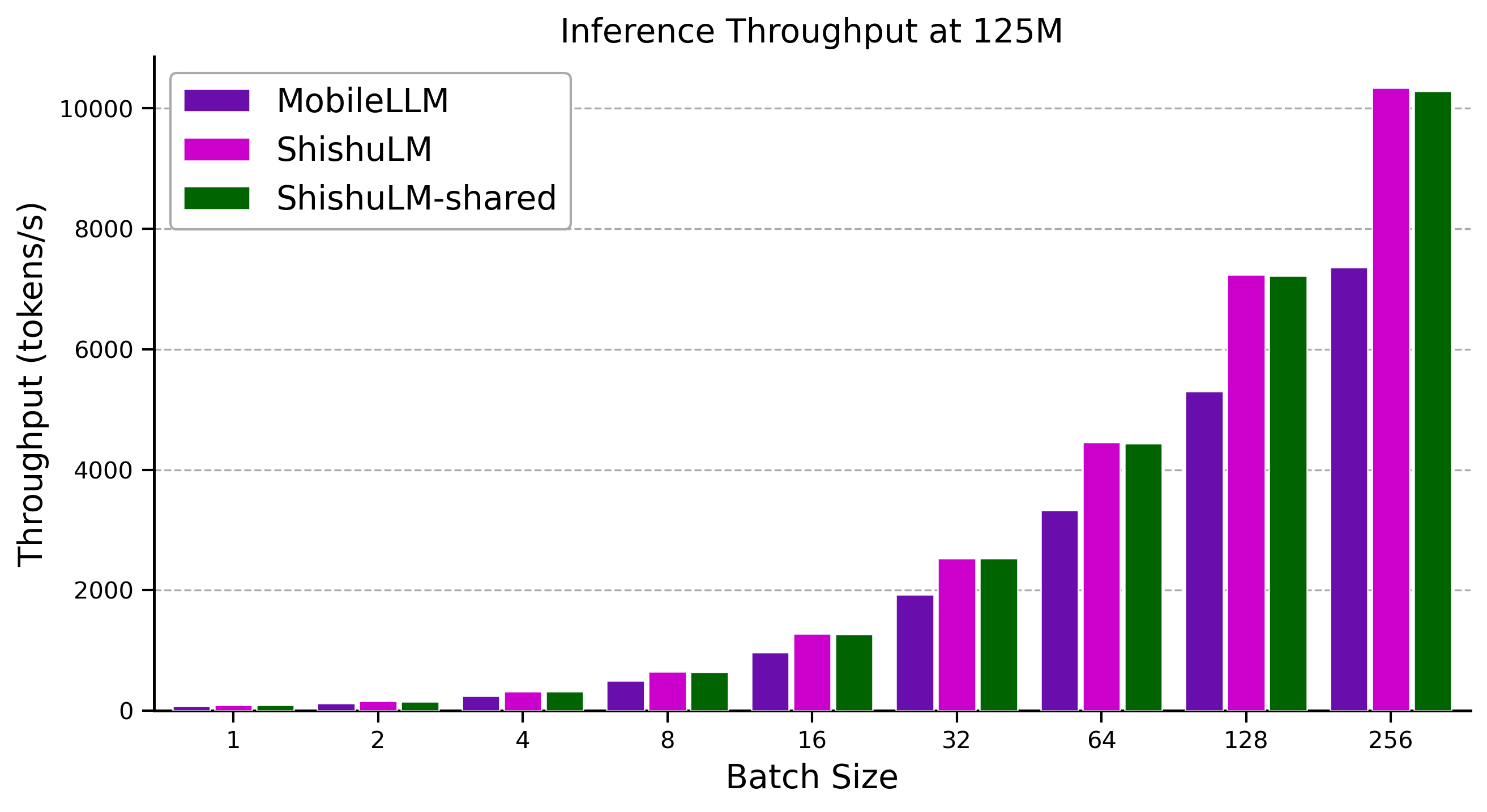}
    \caption{125M}
\end{subfigure}
\begin{subfigure}{0.4\linewidth}
    \includegraphics[width=\linewidth]{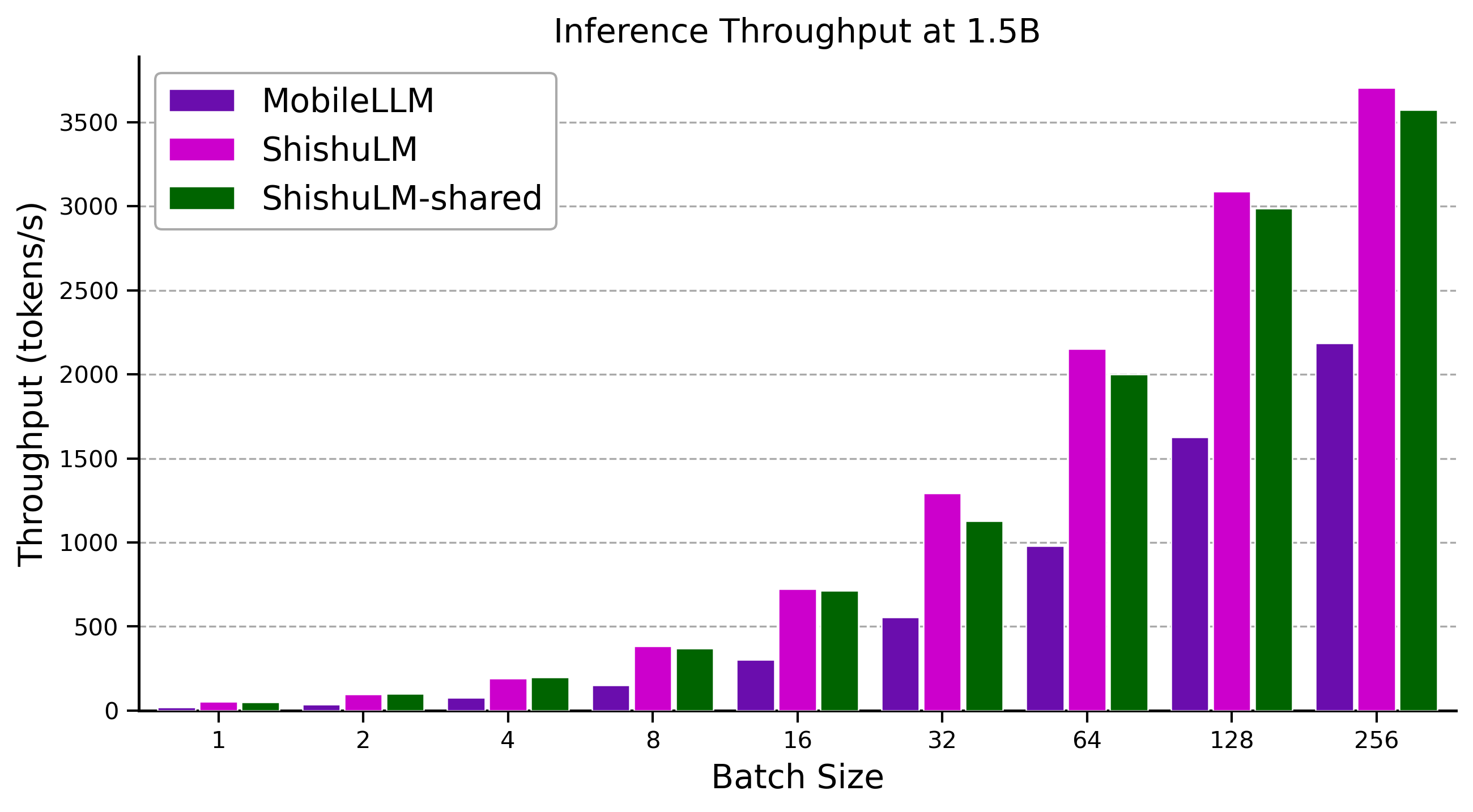}
    \caption{1.5B}
\end{subfigure}

\caption{Inference throughput (tokens/s) of models at the (a) 125M and (b) 1.5B scale, with different batch sizes. Prefill Cache is set to 2048 tokens and the models are allowed to generate 128 tokens.}
\label{fig:throughput}
\end{figure*}

\section{Experiments}
\label{sec:experiments}
In this section, we describe our pre-training experiments. 


\paragraph{Setup.}
Due to their strong performance at smaller scales, we use the MobileLLM model family~\cite{mobilellm}, with parameter counts ranging from 125M to 1.5B, as our base architectures. These models follow the LLaMA  architecture~\cite{touvron2023llamaopenefficientfoundation}. All pre-training experiments are conducted on 8 H200 GPUs using subsets of the SlimPajama dataset~\cite{cerebras2023slimpajama}. Additional training details are provided in Appendix~\ref{sec:training_details}.


\subsection{Pre-training with Fewer Attention Layers}
\label{sec:exp_fewer_attn}
Motivated by the evidence of redundancy in attention computation in the later layers,
we first investigate the impact of removing attention sub-layers from the top layers of 125M and 350M MobileLLM models, in steps of 5 layers. We train these modified models on 5B and 10B tokens, respectively, and measure validation perplexity. We find that naively removing attention sub-layers leads to a degradation in performance, as shown in Figure~\ref{fig:with-without-mlp-small}. 
To regain performance, we attempt a simple corrective strategy: adding MLP sub-layers (with the same dimensions as the original model) towards the top to approximately restore the original parameter count. 

In essence, we still remove attention layers in steps of 5, but also add sufficient number of MLPs so that the parameters are approximately 125M and 350M (same as the parent models). We pre-train these models on the same data for each scale, and these parameter-matched models perform on par with the full-attention baselines as shown in Figure~\ref{fig:with-without-mlp-small}. However, when the fraction of retained attention layers drops below roughly $50\%$, we observe significant performance degradation.

To validate this finding at larger data scales and reduce the effects of stochastic noise in training, we extended training of these 125M and 350M models to 50B tokens, along with the parent models. The same pattern holds (as shown in Figure~\ref{fig:with-mlp-50b}): retaining up to $2/3$ of the attention layers with sufficient compensating MLPs matches or slightly surpasses baseline perplexity and is the most optimal among models with fewer attention parameters. We adopt this $2/3$ attention retention configuration and apply it to the 600M, 1B, and 1.5B MobileLLM models, pre-training each on 50B tokens. Due to compute constraints we were unable to test other configurations on these scales. However, we observe that this configuration achieves performance similar to and in most cases, slightly better than the respective MobileLLM baselines, with the exception of MobileLLM-350M (see Table~\ref{tab:val_perpl_scales}). 
We call these models ``Shishu Language Models'' (abbv. \textsc{ShishuLM}). The architecture is shown in Figure~\ref{fig:architecture}. These findings reflect a more parameter-efficient allocation, with model capacity in later layers entirely directed towards MLP computation. 

\begin{table}
\centering
\setlength{\tabcolsep}{3pt}
\begin{tabular}{lccc}
\toprule
\textbf{Parent model} & \textbf{Baseline}  & \textbf{Shishu} & \textbf{Shishu-S}\\
\midrule
MobileLLM-125 & 14.55 & \textbf{14.44} & 14.67 \\
MobileLLM-350 & \textbf{12.49} & 12.60 & 12.91 \\
MobileLLM-600 & 11.70 & \textbf{11.65} & 12.00 \\
MobileLLM-1b & \textbf{11.03} & \textbf{11.03} & 11.30 \\
MobileLLM-1.5b & 10.67 & \textbf{10.54} & 10.75 \\
\bottomrule
\end{tabular}
\caption{Validation perplexities after pre-training by applying our proposed changes to the base architectures. Lower is better.}
\label{tab:val_perpl_scales}
\end{table}

\subsection{Weight Sharing Across MLP-only Layers}
\label{sec:exp_weight_sharing}

We hypothesize the loss of performance by naive removal of attention to one or both of the following:
\begin{enumerate}
    \item \textbf{Loss of functional modules:} Removal of attention may eliminate computational pathways. 
    \item \textbf{Loss of parameters:} Simply removing layers reduces the total parameter count, diminishing model expressiveness. 
\end{enumerate}

Although adding sufficient number of MLP sub-layers restores performance, it leads to both increment in the number of sub-layers and parameters, as compared to the models with no extra MLPs. It is therefore not clear as to which of the two factors contribute to the results observed in the previous sub-section. In this sub-section we attempt to reduce the parameter count of \textsc{ShishuLM} by sharing weights across the MLP-only layers. Our motivation is twofold. 
First, prior work supports shared weight and repeated computation ~\cite{csordas2025depth, 2024moeut}: larger models appear to ``stretch'' the same functional computations over more layers. Second, we conduct a weight distribution similarity analysis between adjacent MLP-only layers in our trained \textsc{ShishuLM} models. We measure the Earth Mover's Distance (EMD) across all pairs of layers and observe high similarity between the adjacent layers (See Figure~\ref{fig:emd_sim}). We validate these findings using the the Kolmogorov-Smirnov (KS) statistic, which captures the maximum point-wise deviation between two distributions. As shown in Figure~\ref{fig:ks_sim}, after an intermediate layer, where the dissimilarity peaks, adjacent layers have dissimilarity values very close to 0. Please refer to Appendix~\ref{metrics} for details on these metrics. 

Moreover, since MLP sub-layers account for a significant fraction of total model parameters and contribute disproportionately to storage requirements, weight sharing across adjacent MLP-only layers in pairs of two offers parameter reduction while maintaining the similar depth. 
We pre-train weight-shared variants of all \textsc{ShishuLM} models (125M through 1.5B) by tying the parameters of every two consecutive MLP-only layers (including the normalization layer). When trained on 50B tokens, the degradation remains minimal (see Table~\ref{tab:val_perpl_scales}). We refer to these models by ~\textsc{ShishuLM-X-s}, where X indicates the number of parameters in the parent MobileLLM model.

Since the \textsc{ShishuLM-s} models have fewer parameters than the original MobileLLM baselines, this provides additional evidence for the first hypothesis: the performance degradation from naive attention removal is largely attributable to the loss in computational blocks rather than parametric expressibility, and a few extra MLP layers can account for the computation done by the removed attention sub-layers.

\section{Downstream Performance}
\label{sec:downstream}

We evaluate all our pretrained models on a suite of standard zero-shot commonsense reasoning benchmarks using the EleutherAI LM Evaluation Harness~\cite{eval-harness}: HellaSwag~\cite{zellers-etal-2019-hellaswag}, OpenBookQA~\cite{mihaylov-etal-2018-suit}, WinoGrande~\cite{10.1145/3474381}, ARC-Easy and ARC-Challenge~\cite{clark2018thinksolvedquestionanswering}, BoolQ~\cite{clark-etal-2019-boolq}, Social-IQA~\cite{sap-etal-2019-socialiqa} and PIQA~\cite{Bisk_Zellers_Lebras_Gao_Choi_2020} in Table~\ref{tab:zero_shot}. We also report perplexity on the WikiText-2 dataset~\cite{merity2017pointer} along with model parameters in Table~\ref{tab:wikiperp}.

\begin{table}
\centering
\setlength{\tabcolsep}{4pt}
\begin{tabular}{lcc}
\toprule
\textbf{Model}& \textbf{\# Params} &\textbf{ PPL($\downarrow$)} \\
\midrule
MobileLLM-125 & 124.6M & 38.35 \\
\textsc{ShishuLM-125} & 123.7M & 38.32 \\
\textsc{ShishuLM-125-s} & 107.8M & 39.22 \\
\midrule
MobileLLM-350 & 343.3M & 30.59 \\
\textsc{ShishuLM-350} & 342.8M & 30.88 \\
\textsc{ShishuLM-350-s} & 288.8M & 32.47 \\
\midrule
MobileLLM-600 & 603M & 27.92 \\
\textsc{ShishuLM-600} & 603M & 28.14 \\
\textsc{ShishuLM-600-s} & 497M & 29.50\\
\midrule
MobileLLM-1B  & 1B & 25.60 \\
\textsc{ShishuLM-1B}  & 1B & 25.65 \\
\textsc{ShishuLM-1B-s} & 849M & 26.43 \\
\midrule
MobileLLM-1.5B & 1.55B & 24.49\\
\textsc{ShishuLM-1.5B} & 1.55B & 24.42 \\ 
\textsc{ShishuLM-1.5B-s} & 1.3B & 24.68\\
\hline
\end{tabular}
\caption{Wikitext perplexity of various models. Lower is better.}
\label{tab:wikiperp}
\end{table}

\begin{table*}
\centering
\setlength{\tabcolsep}{3pt}
\begin{tabular}{lcccccccccc}
\toprule 
\textbf{Model} & \textbf{HellaSwag} & \textbf{PIQA} & \textbf{Arc-E} & \textbf{Arc-C} & \textbf{WinoGrande} & \textbf{OBQA} & \textbf{SIQA} & \textbf{BoolQ} & \textbf{Avg} \\
\midrule
MobileLLM-125M & \textbf{30.30} & 60.66 & 36.28 & 20.99 & 51.62 & 26.60 & \textbf{36.64} & 37.58 & 37.58 \\
\textsc{ShishuLM-125M} & 29.83 & \textbf{61.37} & 36.07 & 22.87 & \textbf{52.80} & \textbf{28.60} & 36.49 & \textbf{62.08} & \textbf{41.26} \\
\textsc{ShishuLM-125M-s} & 29.64 & 60.01 & \textbf{36.57} & \textbf{22.95} & 50.83 & 27.80 & \textbf{36.64} & 61.35 & 40.72 \\
\midrule
MobileLLM-350M & 33.06 & \textbf{63.66} & 38.22 & 22.27 & 51.14 & 26.60 & 36.80 & \textbf{62.08} & 41.73 \\
\textsc{ShishuLM-350M} & \textbf{33.10} & \textbf{63.66} & \textbf{38.80} & \textbf{22.70} & \textbf{54.30} & 29.00 & \textbf{37.46} & 60.76 & \textbf{42.47} \\
\textsc{ShishuLM-350M-s} & 32.12 & 62.13 & 37.96 & 22.18 & 50.12 & \textbf{29.20} & 37.05 & 62.02 & 41.60 \\
\midrule
MobileLLM-600M & \textbf{35.07} & 63.71 & 39.02 & 23.38 & 51.62 & \textbf{30.00} & 37.36 & \textbf{61.35} & 42.69 \\
\textsc{ShishuLM-600M} & 34.52 & \textbf{63.93} & \textbf{40.28} & 23.29 & \textbf{52.57} & 29.60 & 37.56 & 61.01 & \textbf{42.84}\\
\textsc{ShishuLM-600M-s} & 33.15 & 63.17 & 38.64 & \textbf{25.00} & 51.38 & 29.80 & \textbf{38.08} & 60.89 & 42.51 \\
\midrule
MobileLLM-1B & 36.49 & 66.54 & \textbf{41.12} & \textbf{24.40} & 50.36 & 28.80 & 37.10 & 56.94 & 42.72 \\
\textsc{ShishuLM-1B} & \textbf{36.59} & \textbf{66.76} & \textbf{41.12} & 23.55 & 50.28 & 29.60 & 37.41 & 61.35 & 43.33 \\
\textsc{ShishuLM-1B-s} & 36.01 & 65.67 & 40.82 & 22.70 & \textbf{52.17} & \textbf{30.00} & \textbf{37.92} & \textbf{61.71} & \textbf{43.38} \\
\midrule
MobileLLM-1.5B & 37.57 & \textbf{66.00} & 41.12 & 23.46 & 51.22 & 28.60 & 37.26 & \textbf{61.19} & 43.30 \\
\textsc{ShishLM-1.5B} & \textbf{38.29} & 65.94 & \textbf{42.42} & \textbf{25.09} & 51.62 & 29.60 & 37.21 & 61.04 & \textbf{43.90} \\
\textsc{ShishuLM-1.5B-s} & 37.62 & 65.51 & 41.84 & 24.23 & \textbf{51.85} & \textbf{29.80} & \textbf{38.13} & 59.63 & 43.58 \\
\bottomrule
\end{tabular}
\caption{Zero-shot evaluation results.}
\label{tab:zero_shot}
\end{table*}

\begin{figure}
\centering
    \includegraphics[width=0.8\linewidth]{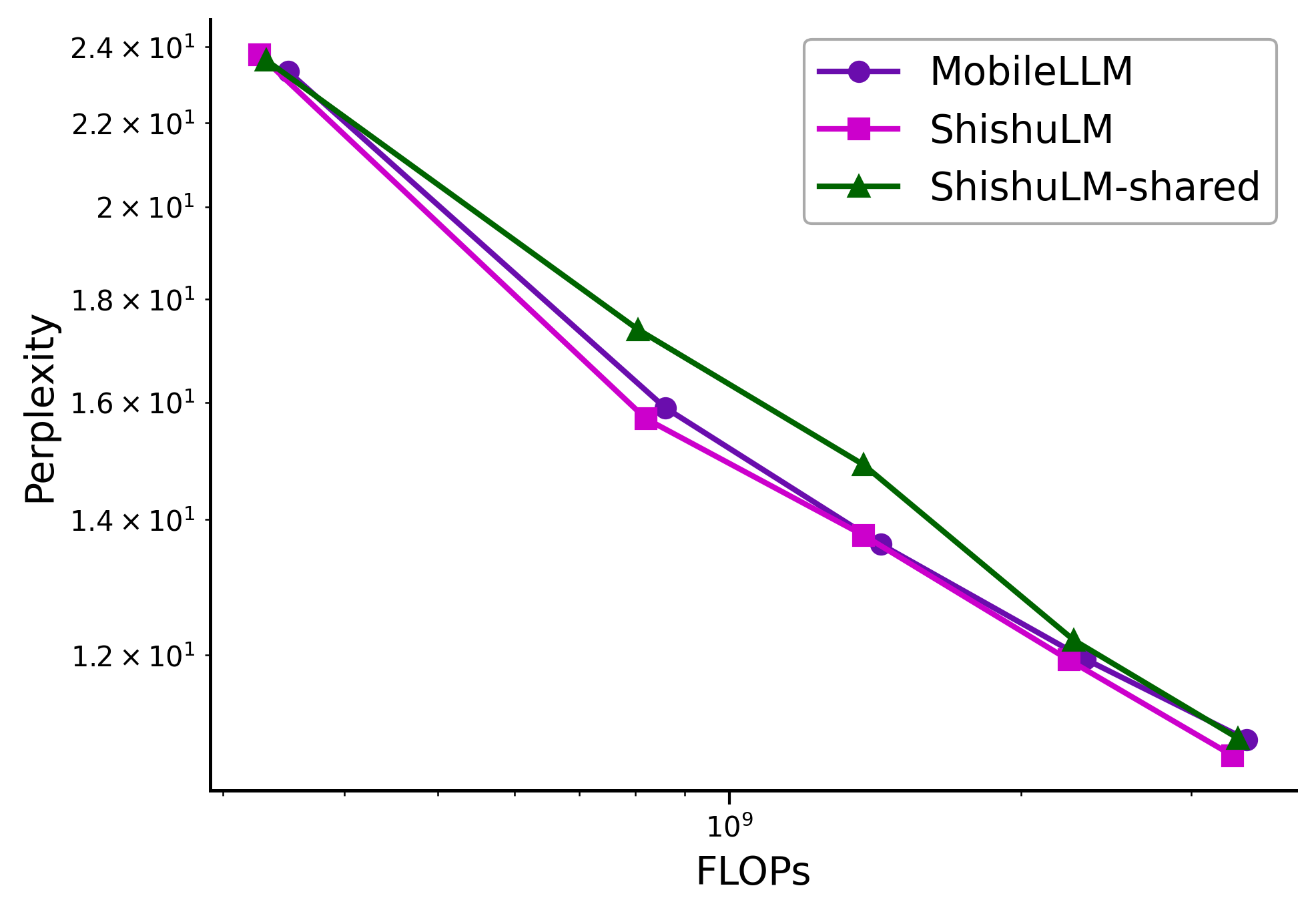}
    \caption{Scaling Laws for our models.}
    \label{fig:scaling}
\end{figure}

\paragraph{Scaling Laws.} Following \citeauthor{gu2024mamba}, we demonstrate Chinchilla scaling \cite{hoffman} for our architectures by training on a compute-optimal budget of 20 tokens per parameter. Our models exhibit scaling behavior similar to the baseline architecture, as illustrated in Figure~\ref{fig:scaling}.

\section{Throughput and Efficiency}
\label{sec:efficiency}
We quantify the efficiency gains achieved by \textsc{ShishuLM} in this section. Since self-attention incurs $\mathcal{O}(T^2)$ computational complexity with respect to sequence length $T$ and MLP operations are $\mathcal{O}(T)$, reducing the number of attention layers leads to tangible throughput improvements. 
In \textsc{ShishuLM}, we remove approximately $1/3$ of full decoder layers and replace them with MLP-only blocks. As the number of added MLP layers is fewer than the number of removed attention sub-layers (approximately 5--7 MLP layers added in place of 10--18 attention sub-layers removed), improvements are seen both at training and inference time.

\paragraph{Training.}
We plot the a single forward pass and a combined forward-backward pass (as relevant for training) across various sequence lengths ranging from 64 to 4096. The results for the 125M and 1.5B scale are shown in Figures~\ref{fig:fwd_latency} and~\ref{fig:fwd_bwd_latency}. Detailed measurement results are reported in Tables~\ref{tab:fwd_latency} and~\ref{tab:fwd_bwd_latency}.

\paragraph{Inference.}
We plot the decrease in generation latency in Figure~\ref{fig:gen_latency} and increase in inference throughput in Figure~\ref{fig:throughput} at the 125M and 1.5B scale. Detailed measurements along can be found in Tables~\ref{tab:generation_latency} and~\ref{tab:throughput}. Memory reduction due to reduction in KV Cache (and reduced parameters in shared versions) are shown in Table~\ref{tab:genetation_memory}. We note that all efficiency improvements reported here are \emph{orthogonal} to complementary optimization techniques such as quantization and efficient attention implementation, and can be coupled with our work. 

\section{Conclusion}

We proposed \textsc{ShishuLM}, an efficient pre-training architecture for decoder-only language models, motivated by depth-wise redundancy in attention sub-layers and the staged nature of information processing in transformers. We demonstrated that replacing the top-layer full decoder blocks with a suitable number of MLP-only blocks, while approximately preserving the total parameter count yields pre-trained models that perform comparably to full-attention baselines across scales from 125M to 1.5B parameters, while achieving meaningful improvements in training and inference throughput. Building on the high distributional similarity between adjacent MLP-only layers, we further introduced a weight-shared variant of \textsc{ShishuLM} that reduces parameter count with minimal performance degradation. 

Our findings are directly grounded in and consistent with interpretability research showing that (i) information integration across tokens is largely complete by the midpoint of the network, (ii) later layers primarily refine the output probability distribution in a manner that does not require cross-token attention, and (iii) adjacent layers perform highly similar computations. We believe our work opens directions for future research into parameter-efficient language model architectures that are grounded in an understanding of how transformers use their computational capacity. 

\section{Limitations}

First, all pre-training experiments are conducted on models up to 1.5B parameters. Second, our experiments are restricted to the LLaMA architectural family. Future work will extend these strategies to other model families such as Qwen~\cite{bai2023qwentechnicalreport}, Gemma~\cite{gemmateam2024}, Pythia~\cite{pythia}, etc., to assess generality. Third, the optimal fraction of retained attention layers (approximately $2/3$ in our setting) may vary with architecture depth, width, or data regime, and a principled approach to determining this ratio would be valuable. Finally, it would be interesting to investigate the reasoning capabilities of post-trained \textsc{ShishuLM} models, and their robustness to known adversarial and privacy attacks in the AI security literature. 

\bibliography{custom}

@inproceedings{attention2017vaswani,
author = {Vaswani, Ashish and Shazeer, Noam and Parmar, Niki and Uszkoreit, Jakob and Jones, Llion and Gomez, Aidan N. and Kaiser, \L{}ukasz and Polosukhin, Illia},
title = {Attention is all you need},
year = {2017},
booktitle = {Proceedings of the 31st International Conference on Neural Information Processing Systems},
pages = {6000–6010},
series = {NIPS'17}
}

@inproceedings{mobilellm,
author = {Liu, Zechun and Zhao, Changsheng and Iandola, Forrest and Lai, Chen and Tian, Yuandong and Fedorov, Igor and Xiong, Yunyang and Chang, Ernie and Shi, Yangyang and Krishnamoorthi, Raghuraman and Lai, Liangzhen and Chandra, Vikas},
title = {MobileLLM: optimizing sub-billion parameter language models for on-device use cases},
year = {2024},
publisher = {JMLR.org},
booktitle = {Proceedings of the 41st International Conference on Machine Learning},
articleno = {1316},
numpages = {24},
location = {Vienna, Austria},
series = {ICML'24}
}

@inproceedings{lad2024infstage,
title={The Remarkable Robustness of {LLM}s: Stages of Inference?},
author={Vedang Lad and Wes Gurnee and Max Tegmark},
booktitle={ICML 2024 Workshop on Mechanistic Interpretability},
year={2024},
url={https://openreview.net/forum?id=R5unwb9KPc}
}

@inproceedings{song2024sleb,
author = {Song, Jiwon and Oh, Kyungseok and Kim, Taesu and Kim, Hyungjun and Kim, Yulhwa and Kim, Jae-Joon},
title = {SLEB: streamlining LLMs through redundancy verification and elimination of transformer blocks},
year = {2024},
publisher = {JMLR.org},
booktitle = {Proceedings of the 41st International Conference on Machine Learning},
articleno = {1876},
numpages = {20},
location = {Vienna, Austria},
series = {ICML'24}
}

@inproceedings{
siddiqui2024deeperlookdepthpruning,
title={A deeper look at depth pruning of {LLM}s},
author={Shoaib Ahmed Siddiqui and Xin Dong and Greg Heinrich and Thomas Breuel and Jan Kautz and David Krueger and Pavlo Molchanov},
booktitle={ICML 2024 Workshop on Theoretical Foundations of Foundation Models},
year={2024},
url={https://openreview.net/forum?id=9B7ayWclwN}
}

@inproceedings{yang2024laco,
    title = "{L}a{C}o: Large Language Model Pruning via Layer Collapse",
    author = "Yang, Yifei  and
      Cao, Zouying  and
      Zhao, Hai",
    editor = "Al-Onaizan, Yaser  and
      Bansal, Mohit  and
      Chen, Yun-Nung",
    booktitle = "Findings of the Association for Computational Linguistics: EMNLP 2024",
    month = nov,
    year = "2024",
    address = "Miami, Florida, USA",
    publisher = "Association for Computational Linguistics",
    url = "https://aclanthology.org/2024.findings-emnlp.372/",
    doi = "10.18653/v1/2024.findings-emnlp.372",
    pages = "6401--6417"
}

@misc{men2024shortgpt,
      title={ShortGPT: Layers in Large Language Models are More Redundant Than You Expect}, 
      author={Xin Men and Mingyu Xu and Qingyu Zhang and Bingning Wang and Hongyu Lin and Yaojie Lu and Xianpei Han and Weipeng Chen},
      year={2024},
      eprint={2403.03853},
      archivePrefix={arXiv},
      primaryClass={cs.CL},
      url={https://arxiv.org/abs/2403.03853}, 
}

@inproceedings{
gu2024mamba,
title={Mamba: Linear-Time Sequence Modeling with Selective State Spaces},
author={Albert Gu and Tri Dao},
booktitle={First Conference on Language Modeling},
year={2024},
url={https://openreview.net/forum?id=tEYskw1VY2}
}

@misc{bai2023qwentechnicalreport,
      title={Qwen Technical Report}, 
      author={Jinze Bai and Shuai Bai and Yunfei Chu and Zeyu Cui and Kai Dang and Xiaodong Deng and Yang Fan and Wenbin Ge and Yu Han and Fei Huang and Binyuan Hui and Luo Ji and Mei Li and Junyang Lin and Runji Lin and Dayiheng Liu and Gao Liu and Chengqiang Lu and Keming Lu and Jianxin Ma and Rui Men and Xingzhang Ren and Xuancheng Ren and Chuanqi Tan and Sinan Tan and Jianhong Tu and Peng Wang and Shijie Wang and Wei Wang and Shengguang Wu and Benfeng Xu and Jin Xu and An Yang and Hao Yang and Jian Yang and Shusheng Yang and Yang Yao and Bowen Yu and Hongyi Yuan and Zheng Yuan and Jianwei Zhang and Xingxuan Zhang and Yichang Zhang and Zhenru Zhang and Chang Zhou and Jingren Zhou and Xiaohuan Zhou and Tianhang Zhu},
      year={2023},
      eprint={2309.16609},
      archivePrefix={arXiv},
      primaryClass={cs.CL},
      url={https://arxiv.org/abs/2309.16609}, 
}

@misc{jiang2023mistral7b,
      title={Mistral 7B}, 
      author={Albert Q. Jiang and Alexandre Sablayrolles and Arthur Mensch and Chris Bamford and Devendra Singh Chaplot and Diego de las Casas and Florian Bressand and Gianna Lengyel and Guillaume Lample and Lucile Saulnier and Lélio Renard Lavaud and Marie-Anne Lachaux and Pierre Stock and Teven Le Scao and Thibaut Lavril and Thomas Wang and Timothée Lacroix and William El Sayed},
      year={2023},
      eprint={2310.06825},
      archivePrefix={arXiv},
      primaryClass={cs.CL},
      url={https://arxiv.org/abs/2310.06825}, 
}

@misc{touvron2023llamaopenefficientfoundation,
      title={LLaMA: Open and Efficient Foundation Language Models}, 
      author={Hugo Touvron and Thibaut Lavril and Gautier Izacard and Xavier Martinet and Marie-Anne Lachaux and Timothée Lacroix and Baptiste Rozière and Naman Goyal and Eric Hambro and Faisal Azhar and Aurelien Rodriguez and Armand Joulin and Edouard Grave and Guillaume Lample},
      year={2023},
      eprint={2302.13971},
      archivePrefix={arXiv},
      primaryClass={cs.CL},
      url={https://arxiv.org/abs/2302.13971}, 
}

@article{javaheripi2023phi,
  title={Phi-2: The surprising power of small language models},
  author={Javaheripi, Mojan and Bubeck, S{\'e}bastien and Abdin, Marah and Aneja, Jyoti and Bubeck, Sebastien and Mendes, Caio C{\'e}sar Teodoro and Chen, Weizhu and Del Giorno, Allie and Eldan, Ronen and Gopi, Sivakanth and others},
  journal={Microsoft Research Blog},
  year={2023}
}

@inproceedings{
gromov2025the,
title={The Unreasonable Ineffectiveness of the Deeper Layers},
author={Andrey Gromov and Kushal Tirumala and Hassan Shapourian and Paolo Glorioso and Dan Roberts},
booktitle={The Thirteenth International Conference on Learning Representations},
year={2025},
url={https://openreview.net/forum?id=ngmEcEer8a}
}

@inproceedings{wu2024layercondensedkvcacheefficient,
    title = "Layer-Condensed {KV} Cache for Efficient Inference of Large Language Models",
    author = "Wu, Haoyi  and
      Tu, Kewei",
    editor = "Ku, Lun-Wei  and
      Martins, Andre  and
      Srikumar, Vivek",
    booktitle = "Proceedings of the 62nd Annual Meeting of the Association for Computational Linguistics (Volume 1: Long Papers)",
    month = aug,
    year = "2024",
    address = "Bangkok, Thailand",
    publisher = "Association for Computational Linguistics",
    url = "https://aclanthology.org/2024.acl-long.602/",
    doi = "10.18653/v1/2024.acl-long.602",
    pages = "11175--11188",
}

@inproceedings{sun2024cacheoncedecoderdecoderarchitectures,
 author = {Sun, Yutao and Dong, Li and Zhu, Yi and Huang, Shaohan and Wang, Wenhui and Ma, Shuming and Zhang, Quanlu and Wang, Jianyong and Wei, Furu},
 booktitle = {Advances in Neural Information Processing Systems},
 editor = {A. Globerson and L. Mackey and D. Belgrave and A. Fan and U. Paquet and J. Tomczak and C. Zhang},
 pages = {7339--7361},
 publisher = {Curran Associates, Inc.},
 title = {You Only Cache Once: Decoder-Decoder Architectures for Language Models},
 url = {https://proceedings.neurips.cc/paper_files/paper/2024/file/0df38cd13520747e1e64e5b123a78ef8-Paper-Conference.pdf},
 volume = {37},
 year = {2024}
}

@inproceedings{rajput2024inferencefriendlymodelsmixattention,
  title = 	 {Inference-Friendly Models With {MixAttention}},
  author =       {Rajput, Shashank and Sheng, Ying and Owen, Sean and Chiley, Vitaliy},
  booktitle = 	 {Proceedings of The 4th NeurIPS Efficient Natural Language and Speech Processing Workshop},
  pages = 	 {370--381},
  year = 	 {2024},
  editor = 	 {Rezagholizadeh, Mehdi and Passban, Peyman and Samiee, Soheila and Partovi Nia, Vahid and Cheng, Yu and Deng, Yue and Liu, Qun and Chen, Boxing},
  volume = 	 {262},
  series = 	 {Proceedings of Machine Learning Research},
  month = 	 {14 Dec},
  publisher =    {PMLR},
  url = 	 {https://proceedings.mlr.press/v262/rajput24a.html},
}

@inproceedings{
ashkboos2024slicegptcompresslargelanguage,
title={Slice{GPT}: Compress Large Language Models by Deleting Rows and Columns},
author={Saleh Ashkboos and Maximilian L. Croci and Marcelo Gennari do Nascimento and Torsten Hoefler and James Hensman},
booktitle={The Twelfth International Conference on Learning Representations},
year={2024},
url={https://openreview.net/forum?id=vXxardq6db}
}

@inproceedings{ma2023llmprunerstructuralpruninglarge,
 author = {Ma, Xinyin and Fang, Gongfan and Wang, Xinchao},
 booktitle = {Advances in Neural Information Processing Systems},
 editor = {A. Oh and T. Naumann and A. Globerson and K. Saenko and M. Hardt and S. Levine},
 pages = {21702--21720},
 publisher = {Curran Associates, Inc.},
 title = {LLM-Pruner: On the Structural Pruning of Large Language Models},
 url = {https://proceedings.neurips.cc/paper_files/paper/2023/file/44956951349095f74492a5471128a7e0-Paper-Conference.pdf},
 volume = {36},
 year = {2023}
}

@inproceedings{frantar2023sparsegptmassivelanguagemodels,
  title = 	 {{S}parse{GPT}: Massive Language Models Can be Accurately Pruned in One-Shot},
  author =       {Frantar, Elias and Alistarh, Dan},
  booktitle = 	 {Proceedings of the 40th International Conference on Machine Learning},
  pages = 	 {10323--10337},
  year = 	 {2023},
  editor = 	 {Krause, Andreas and Brunskill, Emma and Cho, Kyunghyun and Engelhardt, Barbara and Sabato, Sivan and Scarlett, Jonathan},
  volume = 	 {202},
  series = 	 {Proceedings of Machine Learning Research},
  month = 	 {23--29 Jul},
  publisher =    {PMLR},
  url = 	 {https://proceedings.mlr.press/v202/frantar23a.html},
}

@inproceedings{
sun2024simpleeffectivepruningapproach,
title={A Simple and Effective Pruning Approach for Large Language Models},
author={Mingjie Sun and Zhuang Liu and Anna Bair and J Zico Kolter},
booktitle={The Twelfth International Conference on Learning Representations},
year={2024},
url={https://openreview.net/forum?id=PxoFut3dWW}
}

@inproceedings{
brandon2024reducing,
title={Reducing Transformer Key-Value Cache Size with Cross-Layer Attention},
author={William Brandon and Mayank Mishra and Aniruddha Nrusimha and Rameswar Panda and Jonathan Ragan-Kelley},
booktitle={The Thirty-eighth Annual Conference on Neural Information Processing Systems},
year={2024},
url={https://openreview.net/forum?id=M2UzLRoqic}
}

@inproceedings{
loshchilov2018decoupled,
title={Decoupled Weight Decay Regularization},
author={Ilya Loshchilov and Frank Hutter},
booktitle={International Conference on Learning Representations},
year={2019},
url={https://openreview.net/forum?id=Bkg6RiCqY7},
}

@misc{cerebras2023slimpajama,
author = {Soboleva, Daria and Al-Khateeb, Faisal and Myers, Robert and Steeves, Jacob R and Hestness, Joel and Dey, Nolan},
title = {{SlimPajama: A 627B token cleaned and deduplicated version of RedPajama}},
month = June,
year = 2023,
}

@inproceedings{
merity2017pointer,
title={Pointer Sentinel Mixture Models},
author={Stephen Merity and Caiming Xiong and James Bradbury and Richard Socher},
booktitle={International Conference on Learning Representations},
year={2017},
url={https://openreview.net/forum?id=Byj72udxe}
}

@article{ELFWING20183,
title = {Sigmoid-weighted linear units for neural network function approximation in reinforcement learning},
journal = {Neural Networks},
volume = {107},
pages = {3-11},
year = {2018},
note = {Special issue on deep reinforcement learning},
issn = {0893-6080},
doi = {https://doi.org/10.1016/j.neunet.2017.12.012},
url = {https://www.sciencedirect.com/science/article/pii/S0893608017302976},
author = {Stefan Elfwing and Eiji Uchibe and Kenji Doya},
}

@inproceedings{
loshchilov2017sgdr,
title={{SGDR}: Stochastic Gradient Descent with Warm Restarts},
author={Ilya Loshchilov and Frank Hutter},
booktitle={International Conference on Learning Representations},
year={2017},
url={https://openreview.net/forum?id=Skq89Scxx}
}

@misc{eval-harness,
  author       = {Gao, Leo and Tow, Jonathan and Abbasi, Baber and Biderman, Stella and Black, Sid and DiPofi, Anthony and Foster, Charles and Golding, Laurence and Hsu, Jeffrey and Le Noac'h, Alain and Li, Haonan and McDonell, Kyle and Muennighoff, Niklas and Ociepa, Chris and Phang, Jason and Reynolds, Laria and Schoelkopf, Hailey and Skowron, Aviya and Sutawika, Lintang and Tang, Eric and Thite, Anish and Wang, Ben and Wang, Kevin and Zou, Andy},
  title        = {A framework for few-shot language model evaluation},
  month        = 12,
  year         = 2023,
  publisher    = {Zenodo},
  version      = {v0.4.0},
  doi          = {10.5281/zenodo.10256836},
  url          = {https://zenodo.org/records/10256836}
}

@inproceedings{zellers-etal-2019-hellaswag,
    title = "{H}ella{S}wag: Can a Machine Really Finish Your Sentence?",
    author = "Zellers, Rowan  and
      Holtzman, Ari  and
      Bisk, Yonatan  and
      Farhadi, Ali  and
      Choi, Yejin",
    editor = "Korhonen, Anna  and
      Traum, David  and
      M{\`a}rquez, Llu{\'i}s",
    booktitle = "Proceedings of the 57th Annual Meeting of the Association for Computational Linguistics",
    month = jul,
    year = "2019",
    address = "Florence, Italy",
    publisher = "Association for Computational Linguistics",
    url = "https://aclanthology.org/P19-1472/",
    doi = "10.18653/v1/P19-1472",
    pages = "4791--4800",
}

@inproceedings{mihaylov-etal-2018-suit,
    title = "Can a Suit of Armor Conduct Electricity? A New Dataset for Open Book Question Answering",
    author = "Mihaylov, Todor  and
      Clark, Peter  and
      Khot, Tushar  and
      Sabharwal, Ashish",
    editor = "Riloff, Ellen  and
      Chiang, David  and
      Hockenmaier, Julia  and
      Tsujii, Jun{'}ichi",
    booktitle = "Proceedings of the 2018 Conference on Empirical Methods in Natural Language Processing",
    month = oct # "-" # nov,
    year = "2018",
    address = "Brussels, Belgium",
    publisher = "Association for Computational Linguistics",
    url = "https://aclanthology.org/D18-1260/",
    doi = "10.18653/v1/D18-1260",
    pages = "2381--2391",
}

@article{10.1145/3474381,
author = {Sakaguchi, Keisuke and Bras, Ronan Le and Bhagavatula, Chandra and Choi, Yejin},
title = {WinoGrande: an adversarial winograd schema challenge at scale},
year = {2021},
issue_date = {September 2021},
publisher = {Association for Computing Machinery},
address = {New York, NY, USA},
volume = {64},
number = {9},
issn = {0001-0782},
url = {https://doi.org/10.1145/3474381},
doi = {10.1145/3474381},
journal = {Commun. ACM},
month = aug,
pages = {99–106},
numpages = {8}
}

@misc{clark2018thinksolvedquestionanswering,
      title={Think you have Solved Question Answering? Try ARC, the AI2 Reasoning Challenge}, 
      author={Peter Clark and Isaac Cowhey and Oren Etzioni and Tushar Khot and Ashish Sabharwal and Carissa Schoenick and Oyvind Tafjord},
      year={2018},
      eprint={1803.05457},
      archivePrefix={arXiv},
      primaryClass={cs.AI},
      url={https://arxiv.org/abs/1803.05457}, 
}

@inproceedings{clark-etal-2019-boolq,
    title = "{B}ool{Q}: Exploring the Surprising Difficulty of Natural Yes/No Questions",
    author = "Clark, Christopher  and
      Lee, Kenton  and
      Chang, Ming-Wei  and
      Kwiatkowski, Tom  and
      Collins, Michael  and
      Toutanova, Kristina",
    editor = "Burstein, Jill  and
      Doran, Christy  and
      Solorio, Thamar",
    booktitle = "Proceedings of the 2019 Conference of the North {A}merican Chapter of the Association for Computational Linguistics: Human Language Technologies, Volume 1 (Long and Short Papers)",
    month = jun,
    year = "2019",
    address = "Minneapolis, Minnesota",
    publisher = "Association for Computational Linguistics",
    url = "https://aclanthology.org/N19-1300/",
    doi = "10.18653/v1/N19-1300",
    pages = "2924--2936",
}

@article{Bisk_Zellers_Lebras_Gao_Choi_2020, 
title={PIQA: Reasoning about Physical Commonsense in Natural Language}, volume={34}, url={https://ojs.aaai.org/index.php/AAAI/article/view/6239}, DOI={10.1609/aaai.v34i05.6239},
number={05}, 
journal={Proceedings of the AAAI Conference on Artificial Intelligence}, 
author={Bisk, Yonatan and Zellers, Rowan and Le bras, Ronan and Gao, Jianfeng and Choi, Yejin}, 
year={2020}, month={Apr.}, pages={7432-7439} }

@inproceedings{katharopoulos2020transformersarernns,
author = {Katharopoulos, Angelos and Vyas, Apoorv and Pappas, Nikolaos and Fleuret, Fran\c{c}ois},
title = {Transformers are RNNs: fast autoregressive transformers with linear attention},
year = {2020},
publisher = {JMLR.org},
booktitle = {Proceedings of the 37th International Conference on Machine Learning},
articleno = {478},
numpages = {10},
series = {ICML'20}
}

@inproceedings{wolf-etal-2020-transformers,
    title = "Transformers: State-of-the-Art Natural Language Processing",
    author = "Wolf, Thomas  and
      Debut, Lysandre  and
      Sanh, Victor  and
      Chaumond, Julien  and
      Delangue, Clement  and
      Moi, Anthony  and
      Cistac, Pierric  and
      Rault, Tim  and
      Louf, Remi  and
      Funtowicz, Morgan  and
      Davison, Joe  and
      Shleifer, Sam  and
      von Platen, Patrick  and
      Ma, Clara  and
      Jernite, Yacine  and
      Plu, Julien  and
      Xu, Canwen  and
      Le Scao, Teven  and
      Gugger, Sylvain  and
      Drame, Mariama  and
      Lhoest, Quentin  and
      Rush, Alexander",
    editor = "Liu, Qun  and
      Schlangen, David",
    booktitle = "Proceedings of the 2020 Conference on Empirical Methods in Natural Language Processing: System Demonstrations",
    month = oct,
    year = "2020",
    address = "Online",
    publisher = "Association for Computational Linguistics",
    url = "https://aclanthology.org/2020.emnlp-demos.6/",
    doi = "10.18653/v1/2020.emnlp-demos.6",
    pages = "38--45",
}

@inproceedings{
dao2024flashattention,
title={FlashAttention-2: Faster Attention with Better Parallelism and Work Partitioning},
author={Tri Dao},
booktitle={The Twelfth International Conference on Learning Representations},
year={2024},
url={https://openreview.net/forum?id=mZn2Xyh9Ec}
}

@inproceedings{rubner1998earthmover,
  author={Rubner, Y. and Tomasi, C. and Guibas, L.J.},
  booktitle={Sixth International Conference on Computer Vision (IEEE Cat. No.98CH36271)}, 
  title={A metric for distributions with applications to image databases}, 
  year={1998},
  volume={},
  number={},
  pages={59-66},
  doi={10.1109/ICCV.1998.710701}}

@article{su2024roformer,
  title={RoFormer: Enhanced transformer with Rotary Position Embedding},
  author={Su, Jianlin and Lu, Yu and Pan, Shengfeng and Murtadha, Ahmed and Wen, Bo and Liu, Yunfeng},
  journal={Neurocomputing},
  volume={568},
  pages={127063},
  year={2024},
  publisher={Elsevier}
}

@inproceedings{artzy-schwartz-2024-attend,
    title = "Attend First, Consolidate Later: On the Importance of Attention in Different {LLM} Layers",
    author = "Artzy, Amit Ben  and
      Schwartz, Roy",
    editor = "Belinkov, Yonatan  and
      Kim, Najoung  and
      Jumelet, Jaap  and
      Mohebbi, Hosein  and
      Mueller, Aaron  and
      Chen, Hanjie",
    booktitle = "Proceedings of the 7th BlackboxNLP Workshop: Analyzing and Interpreting Neural Networks for NLP",
    month = nov,
    year = "2024",
    address = "Miami, Florida, US",
    publisher = "Association for Computational Linguistics",
    url = "https://aclanthology.org/2024.blackboxnlp-1.10/",
    doi = "10.18653/v1/2024.blackboxnlp-1.10",
    pages = "177--184",
}

@inproceedings{hoffman,
author = {Hoffmann, Jordan and Borgeaud, Sebastian and Mensch, Arthur and Buchatskaya, Elena and Cai, Trevor and Rutherford, Eliza and de Las Casas, Diego and Hendricks, Lisa Anne and Welbl, Johannes and Clark, Aidan and Hennigan, Tom and Noland, Eric and Millican, Katie and van den Driessche, George and Damoc, Bogdan and Guy, Aurelia and Osindero, Simon and Simonyan, Karen and Elsen, Erich and Vinyals, Oriol and Rae, Jack W. and Sifre, Laurent},
title = {Training compute-optimal large language models},
year = {2022},
isbn = {9781713871088},
publisher = {Curran Associates Inc.},
address = {Red Hook, NY, USA},
booktitle = {Proceedings of the 36th International Conference on Neural Information Processing Systems},
articleno = {2176},
numpages = {15},
location = {New Orleans, LA, USA},
series = {NIPS '22}
}

@inproceedings{sap-etal-2019-socialiqa,
  title={Social IQa: Commonsense Reasoning about Social Interactions},
  author={Sap, Maarten and Rashkin, Hannah and Chen, Derek and Le Bras, Ronan and Choi, Yejin},
  booktitle={Proceedings of the 2019 Conference on Empirical Methods in Natural Language Processing and the 9th International Joint Conference on Natural Language Processing (EMNLP-IJCNLP)},
  pages={4463--4473},
  year={2019},
  address={Hong Kong, China},
  publisher={Association for Computational Linguistics},
  url={https://aclanthology.org/D19-1454/},
  doi={10.18653/v1/D19-1454}
}

@article{smirnov1939estimate,
  title={On the estimation of the discrepancy between empirical curves of distribution for two independent samples},
  author={Smirnov, Nikolai V},
  journal={Bulletin Math{\'e}matique de l'Universit{\'e} de Moscou},
  volume={2},
  number={2},
  pages={3--14},
  year={1939}
}

@inproceedings{
2024moeut,
title={Mo{EUT}: Mixture-of-Experts Universal Transformers},
author={R{\'o}bert Csord{\'a}s and Kazuki Irie and J{\"u}rgen Schmidhuber and Christopher Potts and Christopher D Manning},
booktitle={The Thirty-eighth Annual Conference on Neural Information Processing Systems},
year={2024},
url={https://openreview.net/forum?id=ZxVrkm7Bjl}
}

@inproceedings{
csordas2025depth,
title={Do Language Models Use Their Depth Efficiently?},
author={R{\'o}bert Csord{\'a}s and Christopher D Manning and Christopher Potts},
booktitle={The Thirty-ninth Annual Conference on Neural Information Processing Systems},
year={2025},
url={https://openreview.net/forum?id=Kz6eUL86XP}
}

@inproceedings{
lenz2025jamba,
title={Jamba: Hybrid Transformer-Mamba Language Models},
author={Barak Lenz and Opher Lieber and Alan Arazi and Amir Bergman and Avshalom Manevich and Barak Peleg and Ben Aviram and Chen Almagor and Clara Fridman and Dan Padnos and Daniel Gissin and Daniel Jannai and Dor Muhlgay and Dor Zimberg and Edden M. Gerber and Elad Dolev and Eran Krakovsky and Erez Safahi and Erez Schwartz and Gal Cohen and Gal Shachaf and Haim Rozenblum and Hofit Bata and Ido Blass and Inbal Magar and Itay Dalmedigos and Jhonathan Osin and Julie Fadlon and Maria Rozman and Matan Danos and Michael Gokhman and Mor Zusman and Naama Gidron and Nir Ratner and Noam Gat and Noam Rozen and Oded Fried and Ohad Leshno and Omer Antverg and Omri Abend and Or Dagan and Orit Cohavi and Raz Alon and Ro'i Belson and Roi Cohen and Rom Gilad and Roman Glozman and Shahar Lev and Shai Shalev-Shwartz and Shaked Haim Meirom and Tal Delbari and Tal Ness and Tomer Asida and Tom Ben Gal and Tom Braude and Uriya Pumerantz and Josh Cohen and Yonatan Belinkov and Yuval Globerson and Yuval Peleg Levy and Yoav Shoham},
booktitle={The Thirteenth International Conference on Learning Representations},
year={2025},
url={https://openreview.net/forum?id=JFPaD7lpBD}
}

@inproceedings{
queipo2026attention,
title={Attention Sinks and Compression Valleys in {LLM}s are Two Sides of the Same Coin},
author={Enrique Queipo-de-Llano and Alvaro Arroyo and Federico Barbero and Xiaowen Dong and Michael M. Bronstein and Yann LeCun and Ravid Shwartz-Ziv},
booktitle={The Fourteenth International Conference on Learning Representations},
year={2026},
url={https://openreview.net/forum?id=c5TFhCJ6fs}
}

@inproceedings{
barbero2025why,
title={Why do {LLM}s attend to the first token?},
author={Federico Barbero and Alvaro Arroyo and Xiangming Gu and Christos Perivolaropoulos and Petar Veli{\v{c}}kovi{\'c} and Razvan Pascanu and Michael M. Bronstein},
booktitle={Second Conference on Language Modeling},
year={2025},
url={https://openreview.net/forum?id=tu4dFUsW5z}
}

@inproceedings{bodarenko,
 author = {Bondarenko, Yelysei and Nagel, Markus and Blankevoort, Tijmen},
 booktitle = {Advances in Neural Information Processing Systems},
 editor = {A. Oh and T. Naumann and A. Globerson and K. Saenko and M. Hardt and S. Levine},
 pages = {75067--75096},
 publisher = {Curran Associates, Inc.},
 title = {Quantizable Transformers: Removing Outliers by Helping Attention Heads Do Nothing},
 url = {https://proceedings.neurips.cc/paper_files/paper/2023/file/edbcb7583fd8921dad78adecfe06a99b-Paper-Conference.pdf},
 volume = {36},
 year = {2023}
}

@inproceedings{
xiao2024efficient,
title={Efficient Streaming Language Models with Attention Sinks},
author={Guangxuan Xiao and Yuandong Tian and Beidi Chen and Song Han and Mike Lewis},
booktitle={The Twelfth International Conference on Learning Representations},
year={2024},
url={https://openreview.net/forum?id=NG7sS51zVF}
}

@misc{gemmateam2024,
      title={Gemma: Open Models Based on Gemini Research and Technology}, 
      author={Gemma Team and Thomas Mesnard and Cassidy Hardin and Robert Dadashi and Surya Bhupatiraju and Shreya Pathak and Laurent Sifre and Morgane Rivière and Mihir Sanjay Kale and Juliette Love and Pouya Tafti and Léonard Hussenot and Pier Giuseppe Sessa and Aakanksha Chowdhery and Adam Roberts and Aditya Barua and Alex Botev and Alex Castro-Ros and Ambrose Slone and Amélie Héliou and Andrea Tacchetti and Anna Bulanova and Antonia Paterson and Beth Tsai and Bobak Shahriari and Charline Le Lan and Christopher A. Choquette-Choo and Clément Crepy and Daniel Cer and Daphne Ippolito and David Reid and Elena Buchatskaya and Eric Ni and Eric Noland and Geng Yan and George Tucker and George-Christian Muraru and Grigory Rozhdestvenskiy and Henryk Michalewski and Ian Tenney and Ivan Grishchenko and Jacob Austin and James Keeling and Jane Labanowski and Jean-Baptiste Lespiau and Jeff Stanway and Jenny Brennan and Jeremy Chen and Johan Ferret and Justin Chiu and Justin Mao-Jones and Katherine Lee and Kathy Yu and Katie Millican and Lars Lowe Sjoesund and Lisa Lee and Lucas Dixon and Machel Reid and Maciej Mikuła and Mateo Wirth and Michael Sharman and Nikolai Chinaev and Nithum Thain and Olivier Bachem and Oscar Chang and Oscar Wahltinez and Paige Bailey and Paul Michel and Petko Yotov and Rahma Chaabouni and Ramona Comanescu and Reena Jana and Rohan Anil and Ross McIlroy and Ruibo Liu and Ryan Mullins and Samuel L Smith and Sebastian Borgeaud and Sertan Girgin and Sholto Douglas and Shree Pandya and Siamak Shakeri and Soham De and Ted Klimenko and Tom Hennigan and Vlad Feinberg and Wojciech Stokowiec and Yu-hui Chen and Zafarali Ahmed and Zhitao Gong and Tris Warkentin and Ludovic Peran and Minh Giang and Clément Farabet and Oriol Vinyals and Jeff Dean and Koray Kavukcuoglu and Demis Hassabis and Zoubin Ghahramani and Douglas Eck and Joelle Barral and Fernando Pereira and Eli Collins and Armand Joulin and Noah Fiedel and Evan Senter and Alek Andreev and Kathleen Kenealy},
      year={2024},
      eprint={2403.08295},
      archivePrefix={arXiv},
      primaryClass={cs.CL},
      url={https://arxiv.org/abs/2403.08295}, 
}

@inproceedings{pythia,
author = {Biderman, Stella and Schoelkopf, Hailey and Anthony, Quentin and Bradley, Herbie and O'Brien, Kyle and Hallahan, Eric and Khan, Mohammad Aflah and Purohit, Shivanshu and Prashanth, USVSN Sai and Raff, Edward and Skowron, Aviya and Sutawika, Lintang and Van Der Wal, Oskar},
title = {Pythia: a suite for analyzing large language models across training and scaling},
year = {2023},
publisher = {JMLR.org},
booktitle = {Proceedings of the 40th International Conference on Machine Learning},
articleno = {102},
numpages = {34},
location = {Honolulu, Hawaii, USA},
series = {ICML'23}
}

@article{Ku2025SystemsAA,
  title={Systems and Algorithms for Convolutional Multi-Hybrid Language Models at Scale},
  author={Jerome Ku and Eric Nguyen and David W. Romero and Garyk Brixi and Brandon Yang and Anton Vorontsov and Ali Taghibakhshi and Amy X Lu and Dave P. Burke and Greg Brockman and Stefano Massaroli and Christopher R'e and Patrick D. Hsu and Brian L Hie and Stefano Ermon and Michael Poli},
  journal={ArXiv},
  year={2025},
  volume={abs/2503.01868},
  url={https://api.semanticscholar.org/CorpusID:276774804}
}
\appendix

\section{Related Work} 
\label{sec:related_work}
\subsection{Redundancy in LLMs}
Previous work in the field of LLM pruning has identified several redundancies in pre-trained models. At the weight level, \emph{Structured} pruning involves to removing entire rows, columns, or blocks from weight matrices~\cite{ashkboos2024slicegptcompresslargelanguage, ma2023llmprunerstructuralpruninglarge}, while \emph{unstructured} pruning sets individual weights to zero, yielding sparse matrices~\cite{frantar2023sparsegptmassivelanguagemodels, sun2024simpleeffectivepruningapproach}. 
Block-level pruning  eliminates entire decoder layers by quantifying their importance. ~\citet{men2024shortgpt} quantify layer importance using the expected cosine similarity between block inputs and outputs,~\citet{gromov2025the} similarly use cosine similarity to remove contiguous blocks of layers, followed by recovery fine-tuning via QLoRA.~\citet{yang2024laco} replace multiple consecutive layers with a single layer by merging their parameters.~\citet{siddiqui2024deeperlookdepthpruning} propose a Shapley value-based metric over a calibration dataset, while~\citet{song2024sleb} iteratively eliminate the most redundant block based on calibration perplexity. A common finding across these works is that later layers and in particular later-layer attention sub-layers tend to exhibit the highest redundancy. However, all pruning methods continue to bear the full costs of pre-training the original over-parameterised model, while additionally incurring redundancy identification and post-training recovery overheads.

Seeking to reduce KV cache requirements, several works have attempted to share the key-value pairs across attention layers. \citeauthor{brandon2024reducing} proposed sharing KV pairs across adjacent layers in groups of two, reducing the KV cache size by roughly 50\%. \citeauthor{sun2024cacheoncedecoderdecoderarchitectures} also compress the KV cache by 50\% by computing KV pairs only for the first half of the model and sharing them for the remaining layers. \citeauthor{rajput2024inferencefriendlymodelsmixattention} place sliding window attention layers sandwiched between global attention layers, sharing keys and values across the global layers and between consecutive sliding window layers. \citeauthor{wu2024layercondensedkvcacheefficient} further compress the KV cache by computing keys and values only in the top layers and pairing them with queries in the bottom layers.

\subsection{Sub-quadratic Sequence Models and Hybrid Architectures}

A separate line of research attempts to replace the $\mathcal{O}(T^2)$ self-attention mechanism with sub-quadratic alternatives. Linear attention methods~\cite{katharopoulos2020transformersarernns} reformulate attention using kernel approximations to achieve linear-time computation, State space models (SSMs) such as Mamba~\cite{gu2024mamba} sidestep the attention mechanism entirely, and rather than treating all input tokens equally, Mamba dynamically determines which information to retain in its bounded hidden state to achieve $O(1)$ space complexity and $O(T)$ time complexity.

However, these approaches require specialized hardware kernels and benefits are typically obtained at very high context lengths. Hybrid approaches such as models~\cite{lenz2025jamba, Ku2025SystemsAA} interleave attention layers with SSM or local convolution layers, seeking to combine the strengths of both paradigms.

\section{Metrics for measuring similarity in adjacent layer weight distribution} \label{metrics}
We measure the Earth Mover's distance and KS-statistic between the discrete probability distributions of weights in the gate, up and down projection matrices. While the Earth Mover's Distance captures the overall effort required to transform one distribution into another, the KS-statistic captures the maximum point-wise deviation between the distributions. Calculations using both metrics suggest that distributions learned in adjacent layers are very similar.

The formal definitions of these metrics are as follows:

\paragraph{Earth Mover's Distance (EMD)}
Earth Mover's Distance~\cite{rubner1998earthmover} (EMD), also known as the Wasserstein distance, is a metric for quantifying the dissimilarity between probability distributions by framing the comparison as an optimal transport problem. Given two distributions, EMD computes the minimum cost required to transform one distribution into another, where the cost is determined by the amount of \textit{probability mass} that must be moved and the distance over which it must be transported.

In the univariate case, this reduces to computing the integral of the absolute difference of the Cumulative Distribution Functions. Formally, for two univariate distributions $A$ and $B$ with CDFs $F_A(x)$ and $F_B(x)$ respectively, we have
$$EMD(A,B) = \int_{-\infty}^{\infty} | F_A(x) - F_B(x) | \ dx.$$
Since EMD is sensitive to scale, we instead use normalized EMD. For distribution $P_i$ at layer $i$, we have
$$EMD_{norm}(P_i, P_j) = \frac{EMD(P_i, P_j)}{(AAD_i + AAD_j)/2}$$
where
$$AAD_i = \mathbb{E}\left(|P_i - \mathbb{E}(P_i)|\right)$$
is the average absolute deviation from the mean.

\paragraph{Kolmogorov-Smirnov Statistic}
The two-sample Kolmogorov-Smirnov statistic, introduced by~\cite{smirnov1939estimate}, is defined as follows. Let $F_A$ and $F_B$ denote two empirical cumulative distribution functions (ECDFs); then
$$KS(A, B) = \sup_{x \in \mathbb{R}} \left| F_A(x) - F_B(x) \right|$$
where $D \in [0, 1]$, with $D = 0$ indicating identical distributions and $D = 1$ indicating complete separation.

\begin{figure*}[t]
    \centering
    \includegraphics[width=\textwidth]{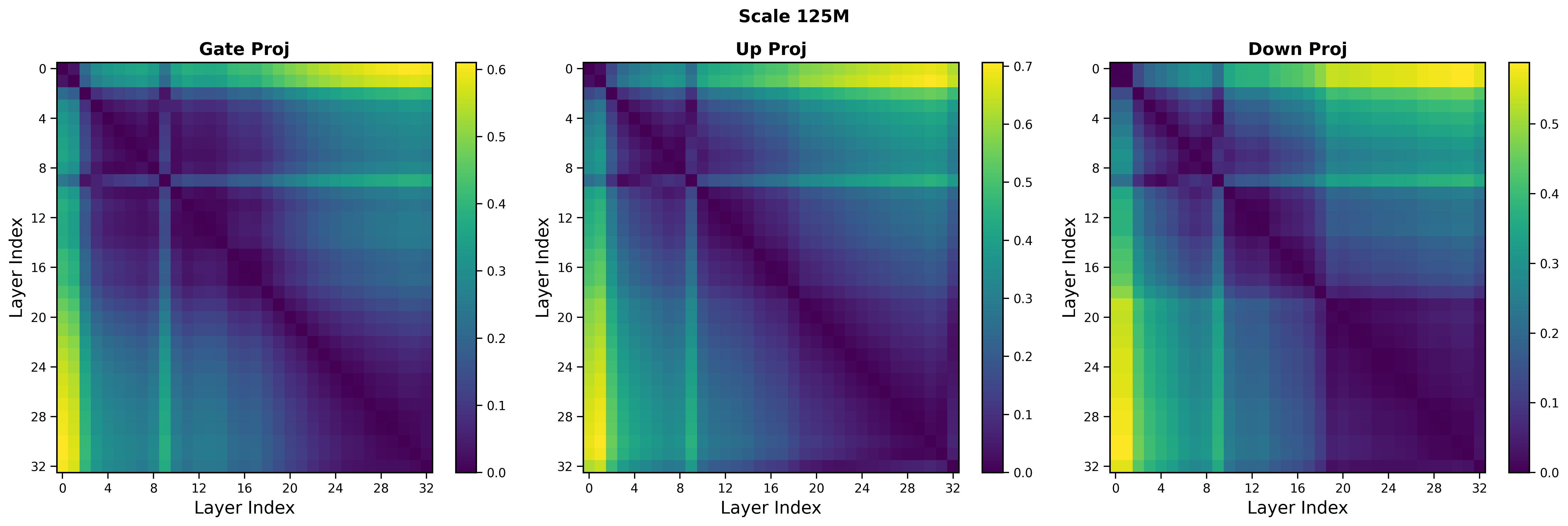}
    \includegraphics[width=\textwidth]{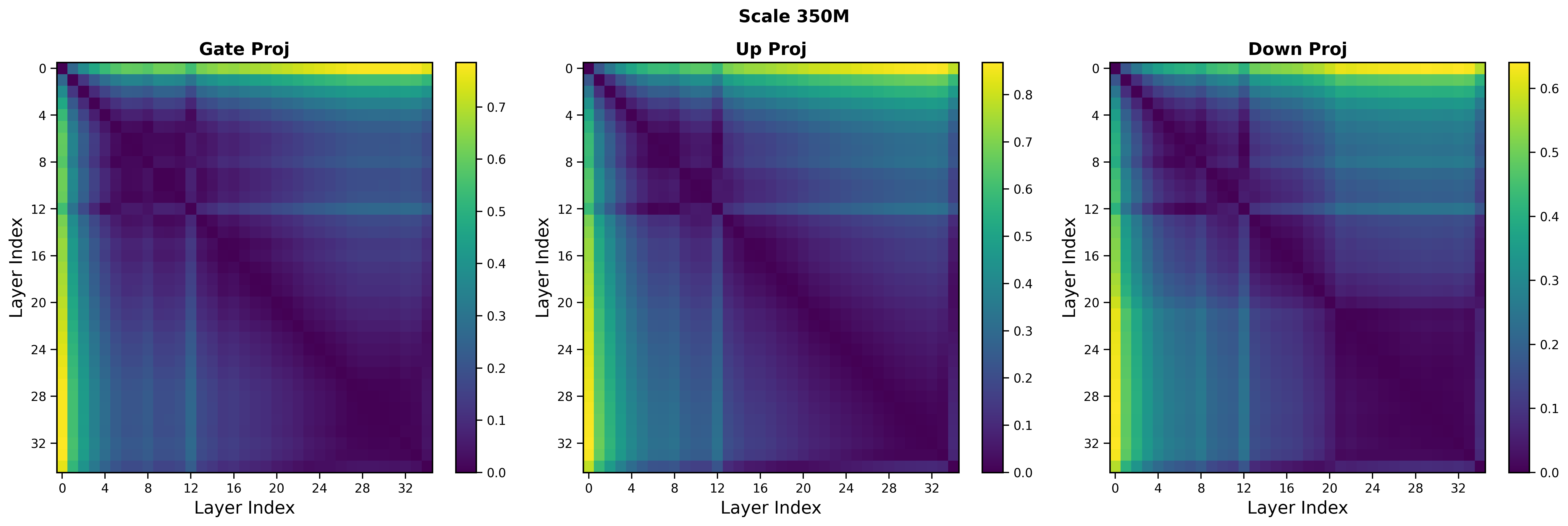}

    \vspace{0.5em}

    \includegraphics[width=\textwidth]{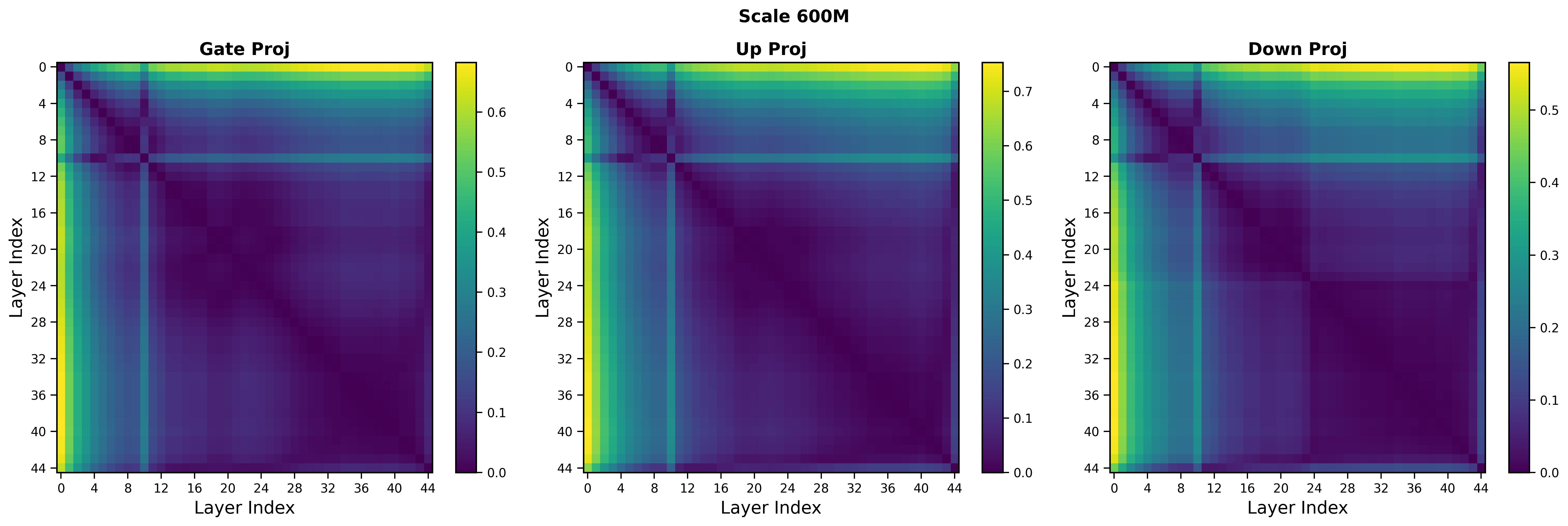}
    \includegraphics[width=\textwidth]{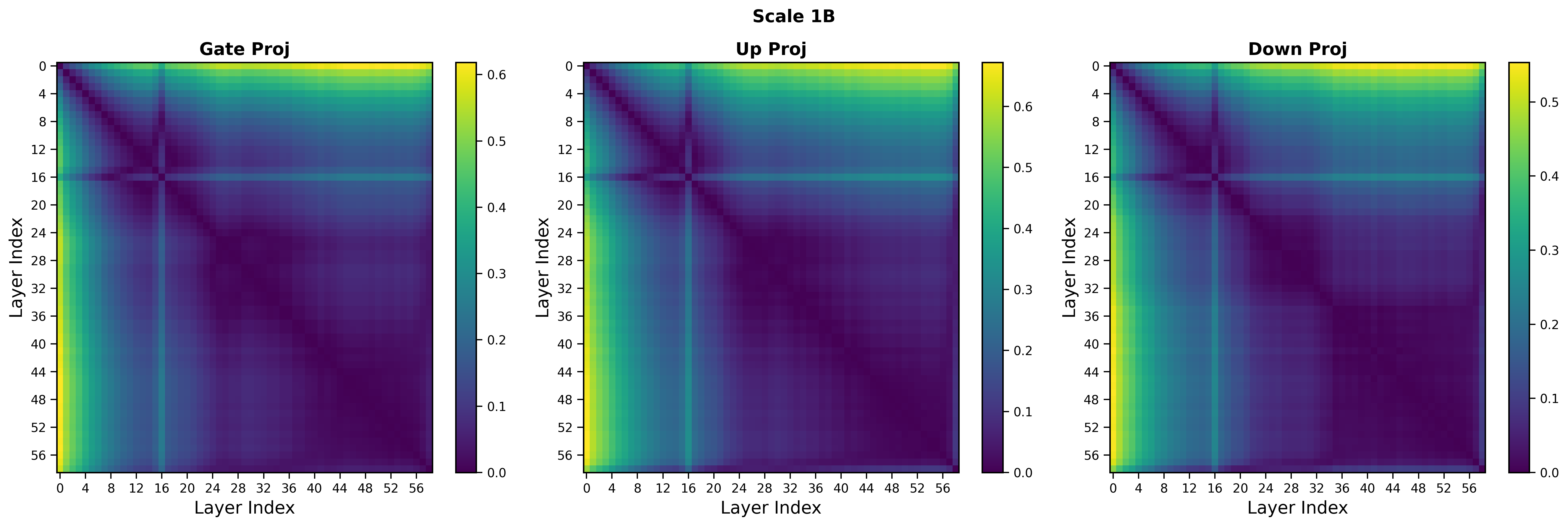}
    \caption{Weight distribution similarity using EMD across all model scales in \textsc{ShishuLM}.}
    \label{fig:emd_sim_all}
\end{figure*}

\begin{figure*}[t]
    \centering
    \includegraphics[width=\textwidth]{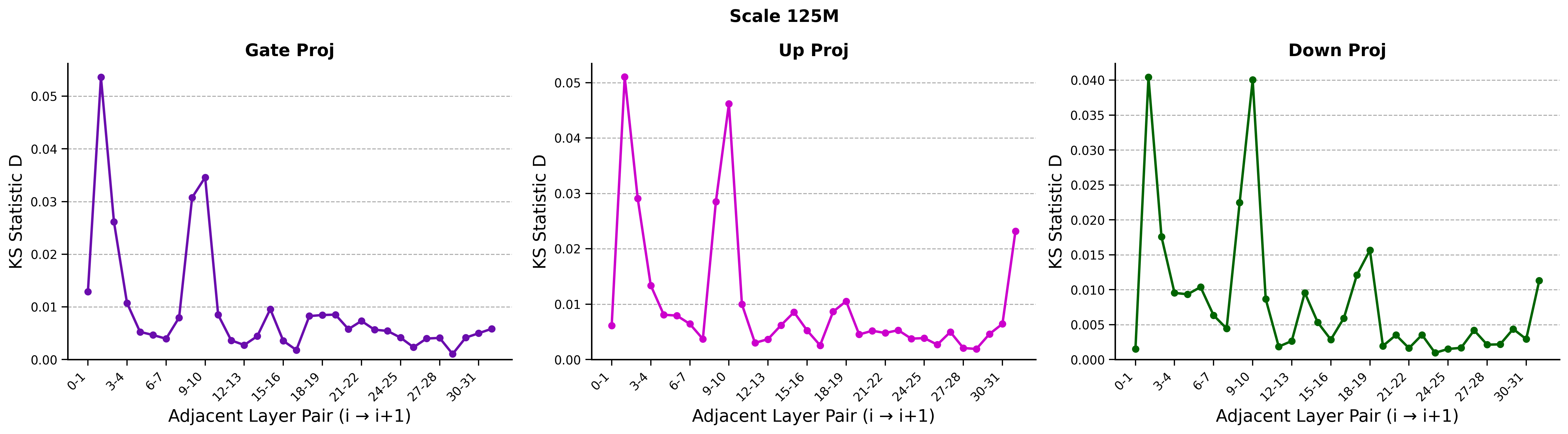}
    \includegraphics[width=\textwidth]{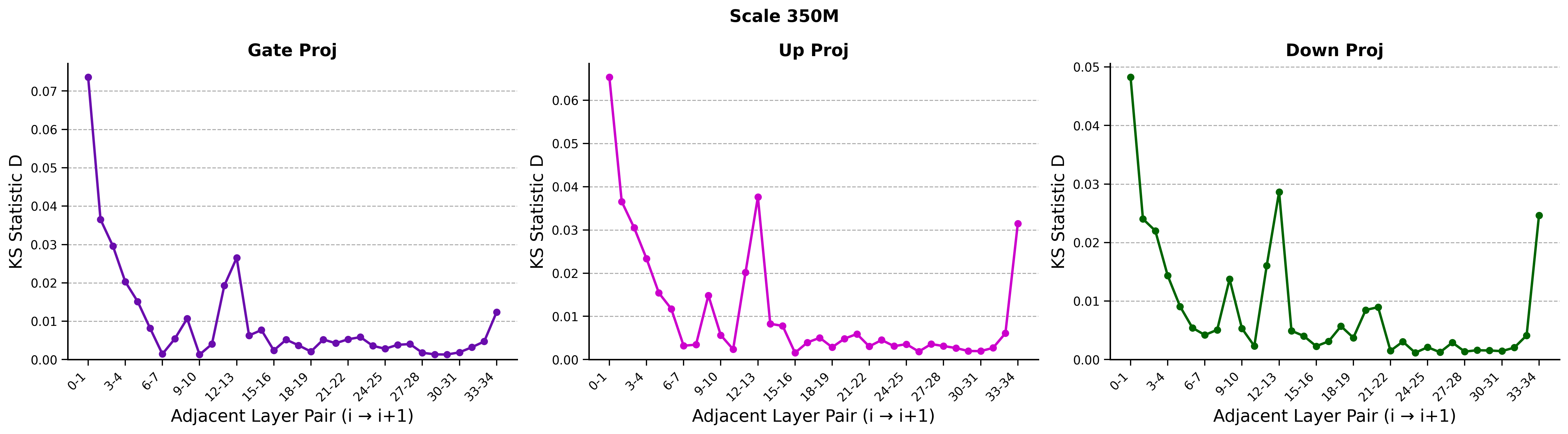}
    \includegraphics[width=\textwidth]{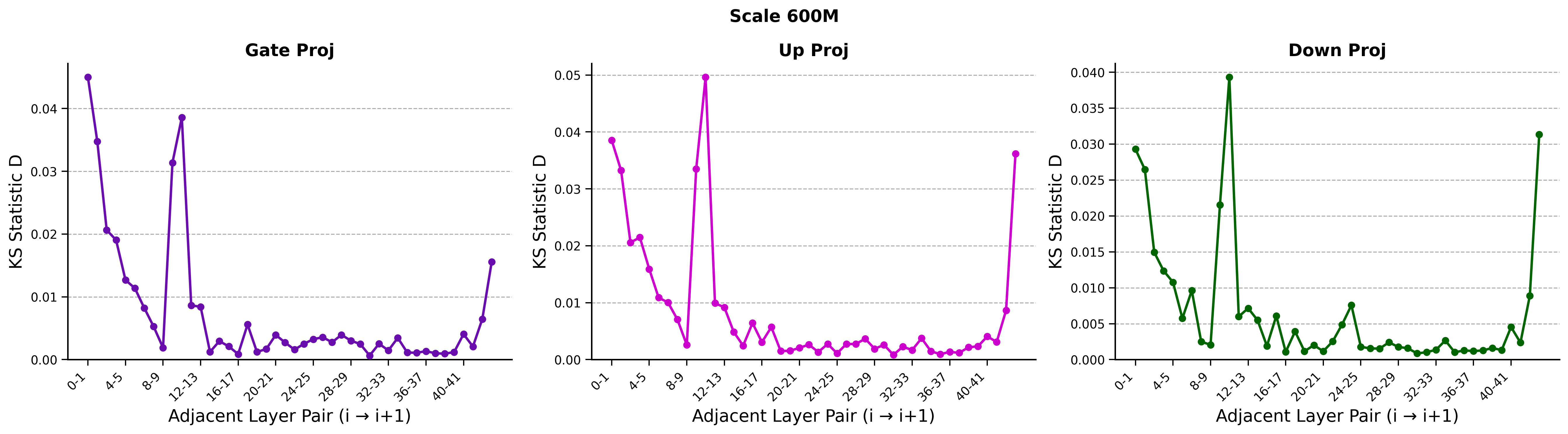}
    \includegraphics[width=\textwidth]{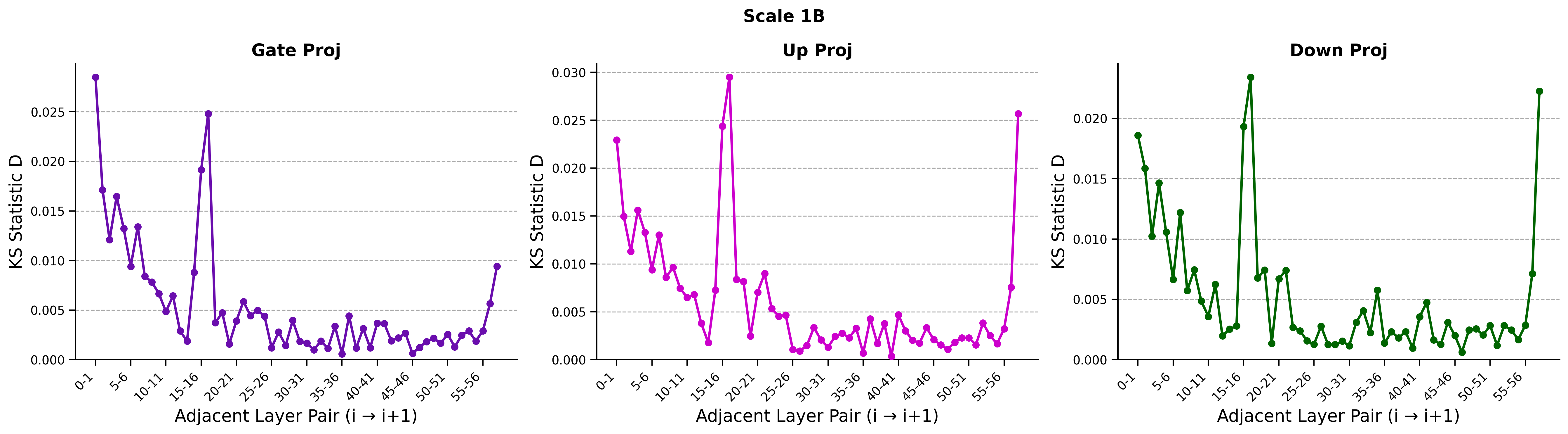}
    \caption{KS-similarity between adjacent layer distributions across all model scales in \textsc{ShishuLM}.}
    \label{fig:ks_sim_all}
\end{figure*}

\section{Training and Architecture details} \label{sec:training_details}

In this section we describe the details of our pre-training experiments. Our implementation utilizes HuggingFace Transformers ~\cite{wolf-etal-2020-transformers} and kernel replacement with FlashAttention 2 ~\cite{dao2024flashattention}. For all our models, we use the Silu activation function~\cite{ELFWING20183} along with rotary position embeddings (RoPE)~\cite{su2024roformer}. 
We use the AdamW optimizer~\cite{loshchilov2018decoupled} with 
\begin{itemize}
    \item $\beta_1 = 0.9$ and $\beta_2 = 0.999$
    \item warmup ratio = $0.05$
    \item weight decay $=1\times10^{-2}$ 
    \item dropout $=0$
\end{itemize}
The learning rates use a cosine scheduler~\cite{loshchilov2017sgdr}. We use gradient clipping with maximum norm $1.0$. RoPE epsilon value is set to $10^{-5}$. The hyperparameters used for the parent models are the same as those for the corresponding derived ones and are described in Table 4.

We use 5, 10 and 50 B token subsets of the SlimPajama dataset~\cite{cerebras2023slimpajama} and tokenize  using the MobileLLM tokenizers.  The block size (sequence length) of a training sample is set to 2048 tokens. Validation perplexity is measured on a hold out set of 2000 sequences of 2048 ($\sim 4$ million tokens). 

\begin{table*}
\centering
\begin{tabular}{lccc}
\toprule
\textbf{Scale} & \textbf{\makecell{Batch-size}} & \textbf{\makecell{Gradient-accumulation}} & \textbf{\makecell{Initial lr}}\\
\midrule
125M &64& 1&$2\times10^{-3}$\\
350M &64& 1 &$2\times10^{-3}$\\
600M &32& 2&$1.5\times10^{-3}$\\
1B & 24 & 3& $1\times10^{-3}$ \\
1.5B & 16  & 3& $1\times10^{-3}$ \\
\bottomrule
\end{tabular}
\label{hyperparameters}
\caption{Hyperparameters using across different scales - per GPU batch size, number of gradient accumulation steps and the initial learning rate.}
\end{table*}

\begin{table*}
\centering
\setlength{\tabcolsep}{4pt}
\begin{tabular}{lcccccc}
\toprule
\textbf{Model} & \textbf{\makecell{Layer\\Config}} & \textbf{\#Head} & \textbf{\#KV-Head} & \textbf{Emb Dim} & \textbf{Hidden Dim} & \textbf{\#Params} \\
\midrule

MobileLLM-125M & 30 + 0 & 9  & 3 & 576  & 1536 & 124,635,456 \\
\textsc{ShishuLM}-125M & 20 + 13 & 9  & 3 & 576  & 1536 & 123,746,688 \\
\textsc{ShishuLM}-125M-shared & 20 + 7*2 & 9  & 3 & 576  & 1536 &  107,817,984\\
\midrule

MobileLLM-350M & 32 + 0 & 15 & 5 & 960  & 2560 & 345,355,200 \\
\textsc{ShishuLM}-350M & 22 + 13 & 15 & 5 & 960  & 2560 & 342,890,880 \\
\textsc{ShishuLM}-350M-shared & 22 + 6*2 & 15 & 5 & 960  & 2560 &  288,816,000\\
\midrule

MobileLLM-600M & 40 + 0& 18 & 6 & 1152 & 3072 & 603,188,352 \\
\textsc{ShishuLM}-600M & 25 + 20 & 18 & 6 & 1152 & 3072 & 603,176,832 \\
\textsc{ShishuLM}-600M-shared & 25 + 10*2 & 18 & 6 & 1152 & 3072 &  496,996,992 \\
\midrule

MobileLLM-1B & 54 + 0& 20 & 5 & 1280 & 3584 & 1,005,461,760 \\
\textsc{ShishuLM}-1B & 36 + 23 & 20 & 5 & 1280 & 3584 & 1,000,529,920 \\
\textsc{ShishuLM}-1B-shared & 36 + 12*2 & 20 & 5 & 1280 & 3584 &  849,127,680\\
\midrule

MobileLLM-1.5B & 54 + 0 & 25 & 5 & 1600 & 4352 & 1,557,844,800 \\
\textsc{ShishuLM}-1.5B & 36 + 23 & 25 & 5 & 1600 & 4352 & 1,556,000,000 \\
\textsc{ShishuLM}-1.5B-shared & 36 + 12*2 & 25 & 5 & 1600 & 4352 &  1,316,692,800\\

\bottomrule
\end{tabular}
\caption{Architecture details of the MobileLLM and \textsc{ShishuLM} variants. Layer config $m + n$ means $m$ full decoder layers followed by $n$ MLP-only layers. In the shared versions we write $m + n*2$ to represent $n$ pairs of weight-shared MLP-layers.}
\label{tab:model_arch}
\end{table*}
All pretraining experiments are done on 8 NVIDIA H200 GPUs with 144 GB VRAM.  We use mixed precision training with bfloat16. The batch sizes are adjusted based on the model size. 

\section{Loss Curves}

\begin{figure*}[t]
    \centering
    \includegraphics[width=0.48\textwidth]{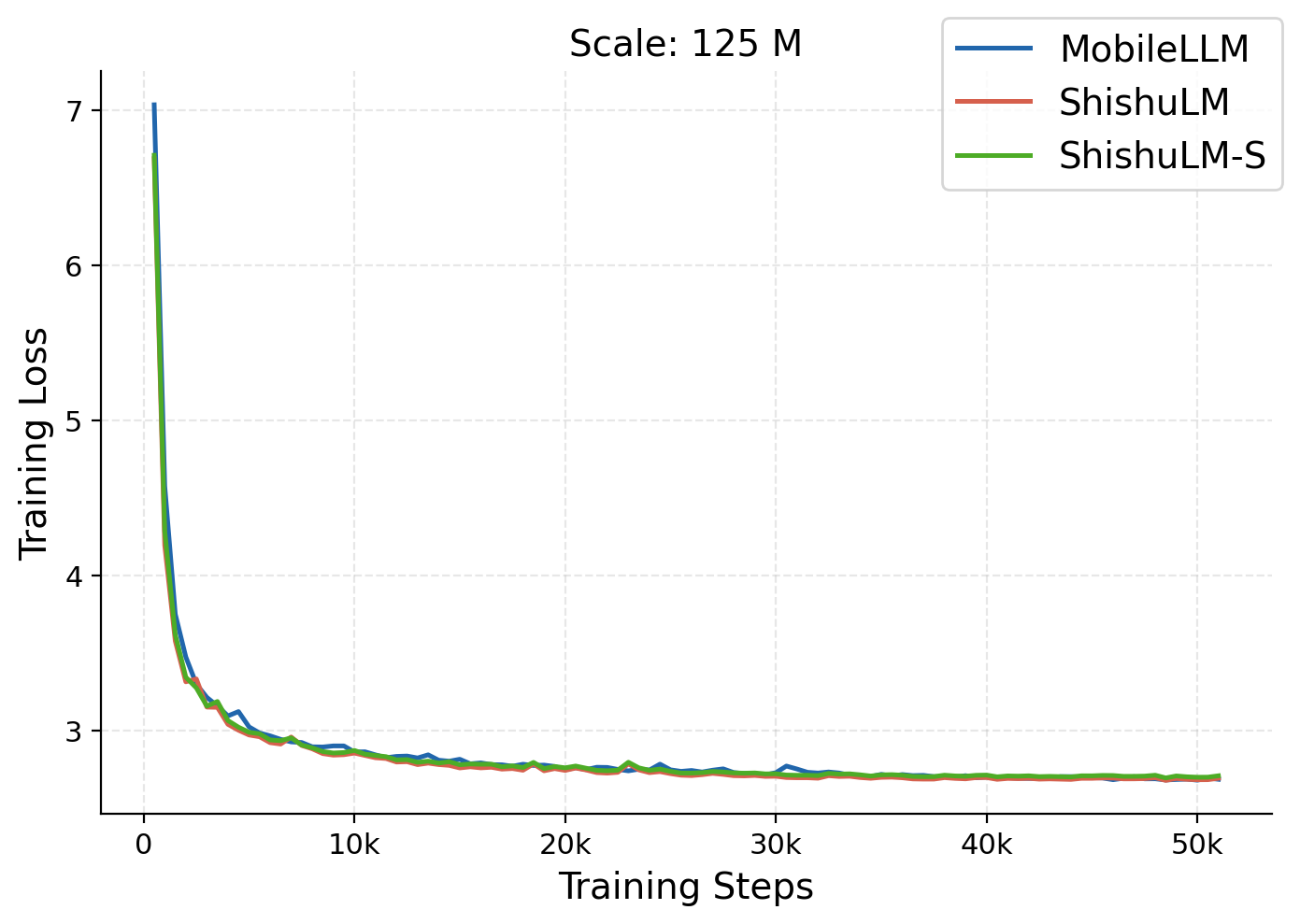}
    \hfill
    \includegraphics[width=0.48\textwidth]{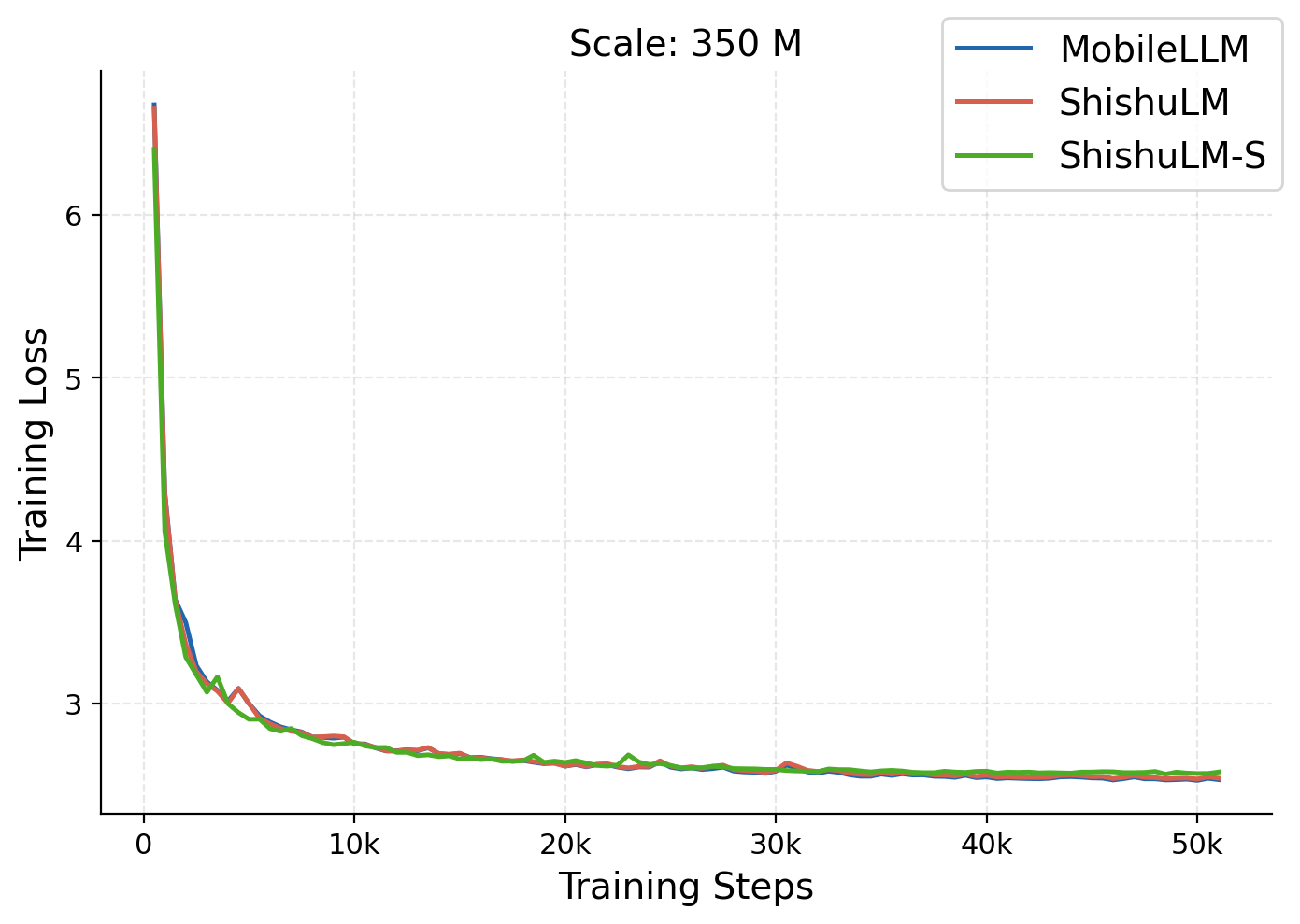}

    \vspace{0.5em}

    \includegraphics[width=0.48\textwidth]{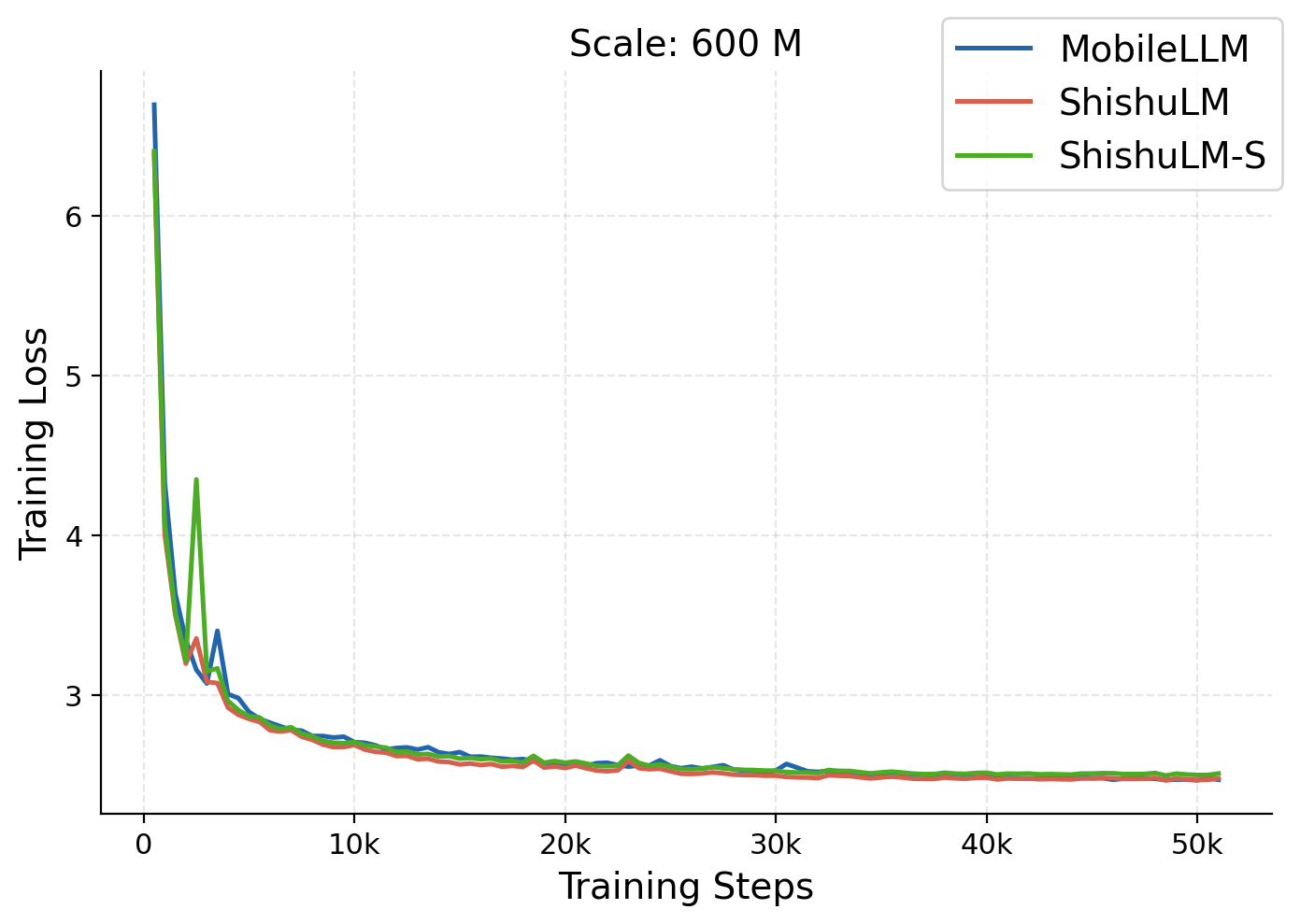}
    \hfill
    \includegraphics[width=0.48\textwidth]{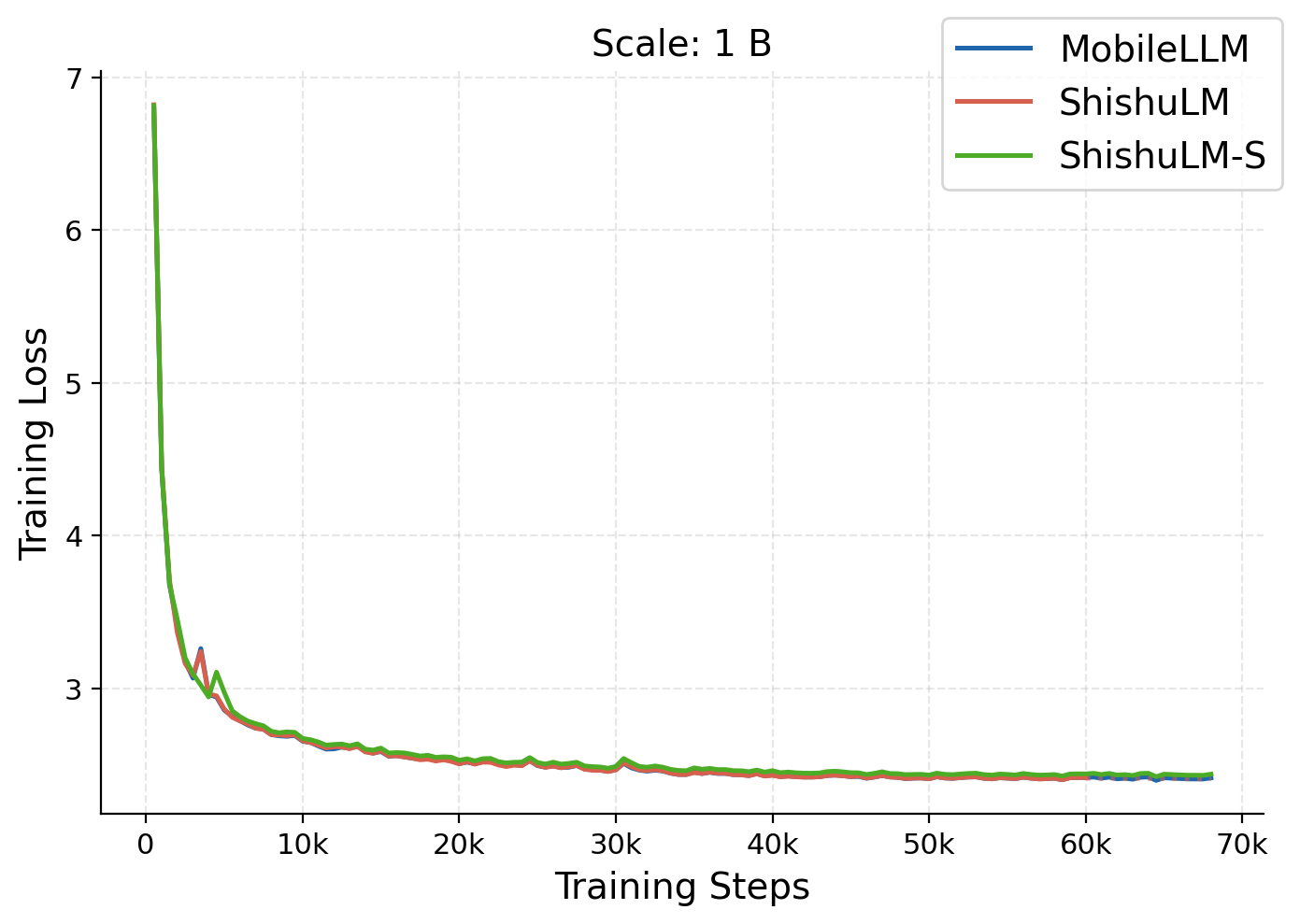}

    \vspace{0.5em}

    \makebox[\textwidth][c]{%
        \includegraphics[width=0.48\textwidth]{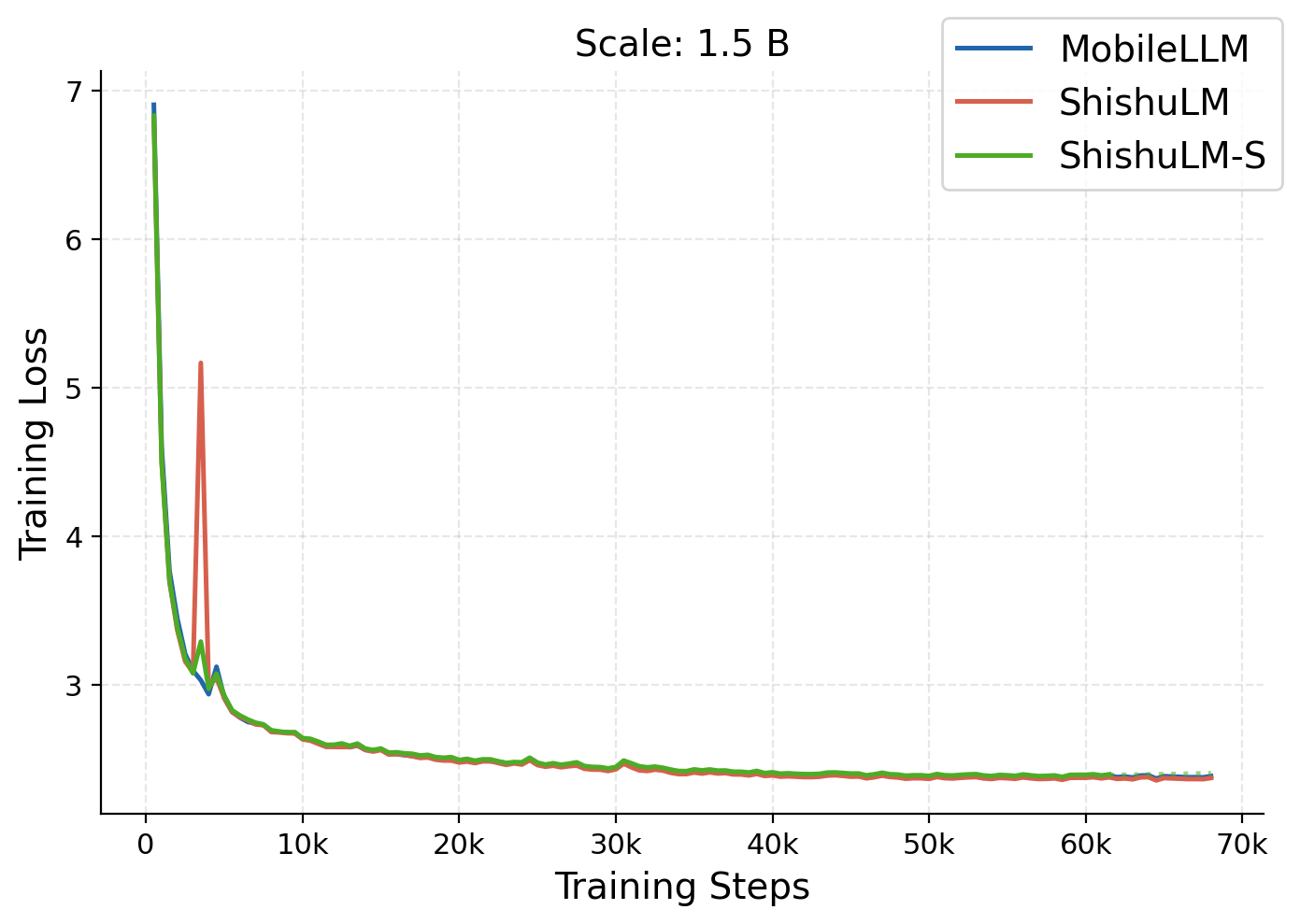}
    }

    \caption{Training loss curves for the five model scales.}
    \label{fig:loss_curves}
\end{figure*}

The plots of training loss against steps is shown in Figure~\ref{fig:loss_curves} for all the models.

\section{Training Speedup}
We measure the forward latency in Table~\ref{tab:fwd_latency} and combined forward + backward latency in ~\ref{tab:fwd_bwd_latency} for all the models across sequence lengths varying from 64 to 4096. 
\begin{table*}
\centering
\begin{tabular}{lccccccc}
\toprule
Model & 64 & 128 & 256 & 512 & 1024 & 2048 & 4096 \\
\midrule
MobileLLM-125 & 33.838 & 33.673 & 34.062 & 35.546 & 36.12 & 36.058 & 38.241 \\
\textsc{ShishuLM-125} & 25.713 & 25.974 & 26.308 & 27.3 & 27.763 & 27.881 & 29.547 \\
\textsc{ShishuLM-125-s} & 26.146 & 26.27 & 26.578 & 27.663 & 28.295 & 28.29 & 30.086 \\
\midrule
MobileLLM-350 & 36.675 & 36.633 & 38.57 & 38.908 & 39.724 & 40.488 & 40.611 \\
\textsc{ShishuLM-350} & 28.873 & 28.811 & 30.48 & 30.863 & 31.241 & 32.213 & 32.12 \\
\textsc{ShishuLM-350-s} & 27.641 & 27.786 & 29.404 & 29.781 & 30.263 & 30.918 & 30.953 \\
\midrule
MobileLLM-600 & 45.905 & 45.968 & 48.257 & 49.156 & 49.949 & 51.14 & 51.175 \\
\textsc{ShishuLM-600} & 18.642 & 18.851 & 21.483 & 32.193 & 21.283 & 20.955 & 26.898 \\
\textsc{ShishuLM-600-s} & 18.183 & 18.424 & 19.481 & 19.381 & 19.689 & 20.376 & 26.53 \\
\midrule
MobileLLM-1B & 42.781 & 42.835 & 46.902 & 46.38 & 62.004 & 63.51 & 63.475 \\
\textsc{ShishuLM-1B} & 24.867 & 25.378 & 27.359 & 27.386 & 26.977 & 27.928 & 39.69 \\
\textsc{ShishuLM-1B-s} & 25.525 & 25.722 & 28.098 & 27.799 & 27.732 & 28.344 & 39.96 \\
\midrule
MobileLLM-1.5B & 60.877 & 64.16 & 64.563 & 46.336 & 47.475 & 59.212 & 96.727 \\
\textsc{ShishuLM-1.5B} & 25.423 & 27.04 & 27.401 & 27.515 & 34.946 & 49.578 & 84.803 \\
\textsc{ShishuLM-1.5B-s} & 25.829 & 27.05 & 27.454 & 27.87 & 35.278 & 50.354 & 85.711 \\
\bottomrule
\end{tabular}
\caption{Forward pass latency (ms) of all the models against sequence length, averaged across 3 trials.}
\label{tab:fwd_latency}
\end{table*}

\begin{table*}
\centering
\begin{tabular}{lccccccc}
\toprule
Model & 64 & 128 & 256 & 512 & 1024 & 2048 & 4096 \\
\midrule
MobileLLM-125 & 82.191 & 82.766 & 83.57 & 86.177 & 88.54 & 89.872 & 93.201 \\
\textsc{ShishuLM-125} & 63.941 & 64.497 & 65.16 & 67.67 & 69.374 & 70.842 & 73.787 \\
\textsc{ShishuLM-125-s} & 65.095 & 65.514 & 66.202 & 68.591 & 70.744 & 71.975 & 75.071 \\
\midrule
MobileLLM-350 & 89.138 & 88.814 & 91.886 & 94.526 & 97.127 & 98.614 & 99.329 \\
\textsc{ShishuLM-350} & 71.373 & 70.759 & 74.009 & 76.056 & 78.287 & 79.633 & 79.606 \\
\textsc{ShishuLM-350-s} & 67.054 & 68.878 & 71.662 & 73.969 & 75.975 & 77.478 & 77.044 \\
\midrule
MobileLLM-600 & 110.907 & 112.194 & 116.174 & 118.92 & 121.432 & 125.012 & 125.321 \\
\textsc{ShishuLM-600} & 45.937 & 46.175 & 78.142 & 74.789 & 49.911 & 52.741 & 80.529 \\
\textsc{ShishuLM-600-s} & 44.739 & 44.94 & 47.058 & 47.433 & 48.373 & 50.521 & 79.434 \\
MobileLLM-1B & 108.693 & 109.072 & 114.121 & 113.433 & 150.873 & 155.192 & 159.653 \\
\textsc{ShishuLM-1B} & 64.178 & 64.591 & 67.812 & 67.319 & 68.135 & 74.648 & 121.447 \\
\textsc{ShishuLM-1B-s} & 63.979 & 64.152 & 67.462 & 67.272 & 67.958 & 75.196 & 122.55 \\
\midrule
MobileLLM-1.5B & 151.583 & 155.057 & 114.899 & 112.472 & 116.32 & 166.525 & 283.634 \\
\textsc{ShishuLM-1.5B} & 64.003 & 65.493 & 66.07 & 64.616 & 91.482 & 143.921 & 259.04 \\
\textsc{ShishuLM-1.5B-s} & 63.672 & 65.448 & 65.949 & 64.901 & 92.488 & 145.469 & 261.981 \\
\bottomrule
\end{tabular}
\caption{Forward + backward latency (ms) of all the models against sequence length, averaged across 3 trials.}
\label{tab:fwd_bwd_latency}
\end{table*}

\section{Inference speedup and memory}
To measure improvements during inference, we set the prefill context to 2048 tokens and measure the time taken to generate 64 to 2048 new tokens in Table~\ref{tab:generation_latency} and the peak GPU memory consumption in~\ref{tab:genetation_memory}. 
Following~\cite{gu2024mamba} we measure the generation throughput by setting the prefill context to 2048 and allowing models to generate 128 tokens for various batch sizes. The throughput is calculated using the formula:
\begin{equation}
\textrm{Throughput} = \frac{127}{\textrm{Total time} - \textrm{TTFT}}
\end{equation}
where TTFT (Time To First Token) represents the time taken to generate the first token. The difference between Total time to generate 128 tokens and TTFT gives the time taken to generate 127 tokens.
The results are shown in Table~\ref{tab:throughput}.

Please note that in all these experiments, the \texttt{$<eos>$} token is set to \texttt{None}, so that the model does not stop generating before the intended number of tokens.
\begin{table*}
\centering
\begin{tabular}{lcccccc}
\toprule
Model & 64 & 128 & 256 & 512 & 1024 & 2048 \\
\midrule
MobileLLM-125 & 1862.351 & 3676.282 & 7387.964 & 14800.752 & 21479.821 & 59027.564 \\
\textsc{ShishuLM-125} & 1405.994 & 2801.33 & 5572.858 & 11140.375 & 22453.405 & 44806.15 \\
\textsc{ShishuLM-125-s} & 1407.675 & 2839.292 & 5686.073 & 11339.453 & 22561.019 & 45125.332 \\
\midrule
MobileLLM-350 & 1975.222 & 3946.171 & 7851.299 & 15700.092 & 26234.292 & 62383.566 \\
\textsc{ShishuLM-350} & 1531.328 & 3064.831 & 6066.522 & 12101.677 & 24315.97 & 48608.603 \\
\textsc{ShishuLM-350-s} & 1481.14 & 2953.442 & 5919.166 & 11772.223 & 23543.782 & 47084.087 \\
\midrule
MobileLLM-600 & 2409.119 & 4357.069 & 9851.366 & 19632.83 & 39126.498 & 78450.316 \\
\textsc{ShishuLM-600} & 912.705 & 1829.799 & 3664.875 & 7503.226 & 14672.568 & 32923.355 \\
\textsc{ShishuLM-600-s} & 905.391 & 1809.028 & 3587.289 & 7179.725 & 14313.755 & 28637.825 \\
MobileLLM-1B & 2584.45 & 5127.022 & 9900.442 & 20585.952 & 52403.289 & 83281.498 \\
\textsc{ShishuLM-1B} & 1260.264 & 2514.476 & 5012.187 & 10003.189 & 19964.959 & 39916.396 \\
\textsc{ShishuLM-1B-s} & 1260.422 & 2519.033 & 5015.874 & 10008.488 & 19985.736 & 39907.113 \\
\midrule
MobileLLM-1.5B & 2235.186 & 4350.952 & 9502.356 & 19049.303 & 37895.821 & 92460.617 \\
\textsc{ShishuLM-1.5B} & 1281.17 & 2523.257 & 5006.245 & 9991.8 & 19934.048 & 39833.26 \\
\textsc{ShishuLM-1.5B-s} & 1289.384 & 2542.772 & 5052.844 & 10084.823 & 20065.022 & 40217.857 \\
\bottomrule
\end{tabular}
\caption{Generation latency (ms) with number of tokens generated, averaged across 3 trials. Prefill cache size is set to 2048 tokens.}
\label{tab:generation_latency}
\end{table*}

\begin{table*}
\centering
\begin{tabular}{lcccccc}
\toprule
Model & 64 & 128 & 256 & 512 & 1024 & 2048 \\
\midrule
MobileLLM-125M & 379.18 & 379.18 & 379.18 & 379.18 & 401.95 & 401.95 \\
\textsc{ShishuLM-125M} & 362.48 & 362.48 & 362.48 & 362.48 & 365.06 & 368.81 \\
\textsc{ShishuLM-125M-s} & 332.41 & 332.41 & 332.41 & 332.41 & 338.04 & 338.04 \\
\midrule
MobileLLM-350M & 890.22 & 890.22 & 890.22 & 890.22 & 892.04 & 926.92 \\
\textsc{ShishuLM-350M} & 854.81 & 854.81 & 854.81 & 854.81 & 854.81 & 866.66 \\
\textsc{ShishuLM-350M-s} & 751.05 & 751.05 & 751.05 & 751.05 & 751.05 & 759.45 \\
\midrule
MobileLLM-600M & 1437.80 & 1437.80 & 1437.80 & 1437.80 & 1438.28 & 1506.06 \\
\textsc{ShishuLM-600M} & 1390.77 & 1390.77 & 1390.77 & 1390.77 & 1390.77 & 1419.29 \\
\textsc{ShishuLM-600M-s} & 1188.24 & 1188.24 & 1188.24 & 1188.24 & 1188.24 & 1216.76 \\
MobileLLM-1B & 2249.00 & 2249.00 & 2249.00 & 2270.00 & 2272.51 & 2300.64 \\
\textsc{ShishuLM-1B} & 2186.71 & 2186.71 & 2186.71 & 2186.71 & 2186.71 & 2201.23 \\
\textsc{ShishuLM-1B-s} & 1898.30 & 1898.30 & 1898.30 & 1898.30 & 1898.30 & 1912.45 \\
\midrule
MobileLLM-1.5B & 3360.15 & 3360.15 & 3360.15 & 3360.15 & 3360.15 & 3419.11 \\
\textsc{ShishuLM-1.5B} & 3308.44 & 3308.44 & 3308.44 & 3308.44 & 3308.44 & 3322.89 \\
\textsc{ShishuLM-1.5B-s} & 2846.40 & 2846.40 & 2846.40 & 2846.40 & 2846.40 & 2860.85 \\
\bottomrule
\end{tabular}
\caption{Peak GPU memory consumption during generation (MB) with number of tokens generated. Prefill cache size is set to 2048 tokens.}
\label{tab:genetation_memory}
\end{table*}

\begin{table*}
\centering
\setlength{\tabcolsep}{4pt}
\begin{tabular}{lccccccccc}
\toprule
Model & 1 & 2 & 4 & 8 & 16 & 32 & 64 & 128 & 256 \\
\midrule
MobileLLM-125 & 70.89 & 119.93 & 242.79 & 493.60 & 966.65 & 1929.95 & 3326.93 & 5305.26 & 7360.57 \\
\textsc{ShishuLM-125} & 91.25 & 155.71 & 315.61 & 645.06 & 1276.00 & 2527.69 & 4458.25 & 7243.68 & 10343.81 \\
\textsc{ShishuLM-125-s} & 90.95 & 154.13 & 315.90 & 642.64 & 1270.39 & 2531.76 & 4434.59 & 7216.79 & 10290.46 \\
\midrule
MobileLLM-350 & 66.86 & 124.17 & 249.43 & 486.62 & 913.24 & 1615.05 & 2632.69 & 3810.32 & 4565.97 \\
\textsc{ShishuLM-350} & 85.31 & 158.39 & 316.64 & 624.09 & 1174.62 & 2113.64 & 3494.86 & 5254.10 & 6371.47 \\
\textsc{ShishuLM-350-s} & 88.06 & 164.11 & 326.34 & 642.24 & 1211.64 & 2187.68 & 3634.63 & 5385.78 & 6634.17 \\
\midrule
MobileLLM-600 & 53.48 & 99.47 & 198.56 & 385.69 & 705.05 & 1231.44 & 1945.67 & 2740.30 & 3125.01 \\
\textsc{ShishuLM-600} & 72.53 & 135.97 & 274.44 & 533.06 & 977.65 & 1741.91 & 2794.68 & 4075.20 & 4787.77 \\
\textsc{ShishuLM-600-s} & 70.82 & 133.29 & 263.13 & 524.98 & 955.88 & 1709.45 & 2749.60 & 4016.67 & 4790.31 \\
\midrule
MobileLLM-1B & 40.05 & 54.23 & 77.12 & 152.43 & 309.97 & 601.52 & 1079.43 & 1811.72 & 2525.73 \\
\textsc{ShishuLM-1B} & 54.73 & 99.16 & 197.43 & 385.69 & 740.87 & 1333.58 & 2163.78 & 3244.12 & 3887.21 \\
\textsc{ShishuLM-1B-s} & 52.75 & 98.32 & 195.20 & 387.75 & 720.57 & 1300.79 & 2162.05 & 3204.29 & 3883.05 \\
\midrule
MobileLLM-1.5B & 19.57 & 37.88 & 75.87 & 150.86 & 302.83 & 557.00 & 978.34 & 1627.16 & 2186.43 \\
\textsc{ShishuLM-1.5B} &  52.81  &  96.33  &  192.79   &     382.35      &  724.43    &   1293.66  &     2152.20    &   3088.31   &    3704.59 \\
\textsc{ShishuLM-1.5B-s}  & 50.06    &  100.14   &     199.95     &   370.32   &     714.35     &  1128.48     &  2000.32   &    2987.64    &   3574.30 \\
\bottomrule
\end{tabular}
\caption{Generation throughput (tokens/s) of models across different batch sizes. Prefill cache size is set to 2048 tokens and the models are allowed to generate 128 tokens.}
\label{tab:throughput}
\end{table*}

\section{AI assistant usage}
Claude Sonnet 4.6 was used to generate the codes for generating plots and measuring statistics.

\end{document}